\documentclass[conference]{IEEEtran}
\IEEEoverridecommandlockouts
\usepackage{cite}
\usepackage{amsmath,amssymb,amsfonts}
\usepackage{algorithmic}
\usepackage{graphicx}
\usepackage{textcomp}
\usepackage{multirow}
\usepackage{xcolor}
\usepackage{xcolor}
\usepackage{hyperref}
\usepackage{longtable, array}
\def\BibTeX{{\rm B\kern-.05em{\sc i\kern-.025em b}\kern-.08em
    T\kern-.1667em\lower.7ex\hbox{E}\kern-.125emX}}
\begin{document}

\title{A Classification of Heterogeneity in Uncrewed Vehicle Swarms and the Effects of Its Inclusion on Overall Swarm Resilience}

\author{%
\parbox[t]{0.48\textwidth}{\centering
\textbf{Abhishek Joshi}\\
Department of Computer Science\\
Texas A\&M University--Corpus Christi\\
Corpus Christi, Texas 78412\\
ajoshi5@islander.tamucc.edu
}\hfill
\parbox[t]{0.48\textwidth}{\centering
\textbf{Abhishek Phadke}\\
School of Engineering and Computing\\
Christopher Newport University\\
Newport News, Virginia 23608\\
abhishek.phadke@cnu.edu
}\\[0.8em]
\parbox[t]{0.48\textwidth}{\centering
\textbf{Tianxing Chu}\\
Conrad Blucher Institute for Surveying and Science\\
Texas A\&M University--Corpus Christi\\
Corpus Christi, Texas 78412\\
tianxing.chu@tamucc.edu
}\hfill
\parbox[t]{0.48\textwidth}{\centering
\textbf{F. Antonio Medrano}\\
Department of Computer Science\\
Texas A\&M University--Corpus Christi\\
Corpus Christi, Texas 78412\\
antonio.medrano@tamucc.edu
}%
}

\maketitle

\begin{abstract}
Combining different types of agents in uncrewed vehicle (UV) swarms has emerged as an approach to enhance mission resilience and operational capabilities across a wide range of applications. This study offers a systematic framework for grouping different types of swarms based on three main factors:  agent nature (behavior and function), hardware 
structure (physical configuration and sensing capabilities), and 
operational space (domain of operation). A literature review indicates that strategic heterogeneity significantly improves swarm performance.  Operational challenges, including communication architecture constraints, energy-aware coordination strategies, and control system integration, are also discussed. The analysis shows that heterogeneous swarms are more resilient because they can leverage diverse capabilities, adapt roles on the fly, and integrate data from multidimensional sensor feeds. Some important factors to consider when implementing are sim-to-real-world transfer for learned policies, standardized evaluation metrics, and control architectures that can work together. Learning-based coordination, GPS (Global Positioning System)-denied multi-robot SLAM (Simultaneous Localization and Mapping), and domain-specific commercial deployments collectively demonstrate that heterogeneous swarm technology is moving closer to readiness for high-value applications. This study offers a single taxonomy and evidence-based observations on methods for designing mission-ready heterogeneous swarms that balance complexity and increased capability.
\end{abstract}
\begin{IEEEkeywords}
uncrewed aerial vehicles, uncrewed ground vehicles, heterogeneity, resilience
\end{IEEEkeywords}

\section{Introduction}
The development and use of uncrewed vehicles (UVs) has accelerated over the past decade. Often, these vehicles are used to conduct tasks in environments deemed too risky for human operators to work in situ. Since UVs are controlled remotely, the lack of pilots also reduces their size and costs compared to human-crewed vehicles. As remote operation and varying degrees of autonomy are developed, their use is predicted to increase significantly in the coming years. Over time, researchers realized that using multiple agents to form coordinated swarms can accomplish tasks more efficiently. However, with the dynamic operating environment of such swarms and the added complexity of coordinating multiple agents, building resilient swarms has become increasingly complex \cite{drones6110340}. UVs have been developed for almost all types of Earth geography, with advances expanding their use even in outer space \cite{drones6010004}. A combination of such agents within a single swarm topology has created heterogeneous teams capable of functioning as a unit. The swarm's varied agents bring their strengths to the table and complement one another. Such compounded swarms have demonstrated greater resilience than homogeneous (single-type) agent swarms across multiple applications \cite{ir.2023.27}.

Swarm heterogeneity is incorporated into agents through a variety of features discussed in this study. This study classifies the current work on heterogeneous swarm deployments. Multiple studies, such as \cite{Makkapati2020ApolloniusAA, deng2013cooperative, 1501630}, have compared mission progress between homogeneous and heterogeneous swarms, and, when implemented appropriately, have reported improved performance with heterogeneous agents. The measurement of this performance increase can vary, such as a reduction in the time the swarm takes to accomplish a goal, an ability to complete the mission successfully under more unexpected circumstances, or another metric used to monitor swarm performance for that application.

\begin{figure}
    \centering
    \includegraphics[width=0.8\linewidth]{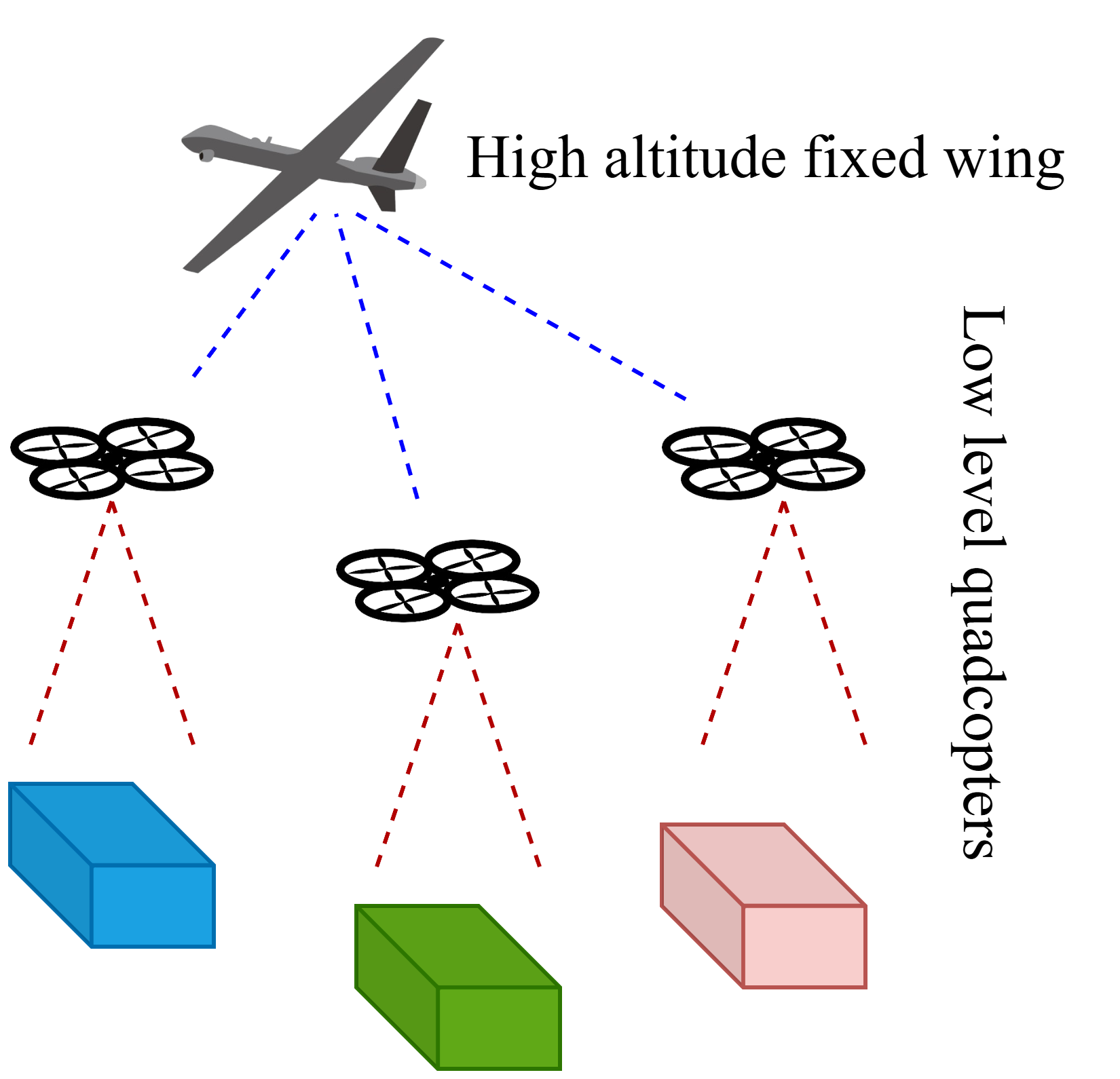}
    \caption{A high altitude fixed wing aircraft forms and maintains a communication link with a low level quadcopter swarm that performs closed range inspection}
    \label{fig:hetero-swarm}
\end{figure}

As mission environments become more uncertain, adversarial, and information-dense, it is vital to study heterogeneity in UV swarms. Homogeneous swarms, although easier to coordinate and model, may lack the diversity needed to respond to real-time failures, domain-specific constraints, or multi-objective mission requirements. When a swarm comprises agents with identical physical capabilities and behavior models, any environmental change or failure mode that affects one agent is likely to affect all others in the same way. This creates brittle swarm architectures with limited resilience under unexpected conditions. In contrast, heterogeneous swarms offer a broader range of responses and allow for intentional distribution of capabilities across multiple roles or platforms. This enhances redundancy and adaptability. For instance, high-endurance aerial platforms can serve as communication relays, while agile rotor-based (quad, hexa, octo) platforms perform close-range inspection. Figure \ref{fig:hetero-swarm} is an illustration of this scenario.

Similarly, introducing land- or water-surface vehicles enables terrain-dependent logistics such as recharging, data offloading, or payload exchange.
There have been some novel implementations of heterogeneous coordination in situations such as underground searches~\cite{tranzatto2022cerberusautonomousleggedaerial}, agriculture automation \cite{JU2022107336}, and GPS blackout scenarios ~\cite{chang2022lamp}. There is a need for an organizing framework for studying heterogeneity across mission spaces. The current literature indicates that heterogeneous swarms can be used in various ways, but no single system organizes all types of heterogeneity. In fact, informal discussions about heterogeneity make it difficult to compare papers, argue for performance gains, and study heterogeneity across design systems for a new mission. A uniform characterization framework helps in understanding the various types of heterogeneity and their external impacts on a swarm's performance, autonomy, and capabilities. Using this framework, researchers will also be able to easily articulate the heterogeneity of their systems and justify their design strategy in line with the mission's requirements. Recent research, namely decentralized coordination ~\cite{goarin2024graph} and asynchronous learning~\cite{10301527}, addresses heterogeneity through different characterizations, including the communication topology and temporal execution. Such studies could benefit from having definitive labels for the heterogeneous characteristics in their swarms. These labels can then be linked to their resilience claims, enabling effective comparisons between labels, and allowing a greater understanding of resilience\\

Through a structured categorization of UV swarms and an investigation of how heterogeneity enhances resilience, this research seeks to complement the idea of resilience through heterogeneity by identifying the underlying factors that distinguish agents.
. From SLAM-enabled multi-robot systems~\cite{chang2022lamp}, autonomous subterranean exploration~\cite{tranzatto2022cerberusautonomousleggedaerial} to learning based coordination ~\cite{goarin2024graph,10301527}, and domain-specific applications like agriculture~\cite{JU2022107336}, this study references recent methods and traces their development on earlier work such as cooperative task allocation~\cite{deng2013cooperative} and swarm coordination~\cite{1501630}.

The remainder of this paper is arranged as follows. Section \ref{equations} presents condensed and formal models for UAV swarms and integration of heterogeneity. 
Section \ref{Classify} provides a taxonomy of heterogeneous UAV swarms, labeling them by operation space, nature, and hardware. Section \ref{op} describes the effects and considerations of heterogeneity on the operation of swarm systems, and Section \ref{overview} (Table \ref{tab:swarm_studies}) describes current work on heterogeneity and assigns labels to the heterogeneity types created in Section \ref{Classify}.
Section\ref{map} provides a heterogeneity to resilience map and  \ref{conclude} has concluding statements.

\section{Condensed formal models for UAV agents, swarms, heterogeneity, and resilience}
\label{equations}
% ---  ---
\subsection{General swarm and agent model}
A heterogeneous uncrewed vehicle swarm can be modeled as a set of agents

\begin{equation}
\mathcal{A} = \{1,2,\dots,N\}
\label{eq:swarm_modeled}
\end{equation}

where each agent \(i \in \mathcal{A}\) has a state vector \(x_i(t)\), control input \(u_i(t)\), 
and dynamics

\begin{equation}
\dot{x}_i(t) = f_i\big(x_i(t), u_i(t), w_i(t)\big),
\label{eq:state_vector}
\end{equation}

with \(w_i(t)\) representing environmental or process disturbances. 
A time-varying graph captures the interaction structure between agents.
\begin{equation}
\mathcal{G}(t) = \big(\mathcal{A}, \mathcal{E}(t)\big),
\label{eq:graph_capture}
\end{equation}

where \(\mathcal{E}(t)\) is the set of sensing or communication edges at time \(t\).
Heterogeneity arises when the functions \(f_i\), the admissible control sets \(\mathcal{U}_i\),
or sensing/actuation capabilities differ across agents.

\subsection{Formalizing Heterogeneity}

% ---  ---
Each agent \(i\) is associated with a feature vector
\begin{equation}
\phi_i = \big(\phi_i^{(N)}, \phi_i^{(H)}, \phi_i^{(O)}\big),
\label{eq:ft_vector}
\end{equation}

where 
\(\phi_i^{(N)}\) encodes nature-based (behavioral) attributes,
\(\phi_i^{(H)}\) encodes hardware/structural attributes,
and \(\phi_i^{(O)}\) encodes operational-space attributes.

A heterogeneity measure for attribute type 
\(\ast \in \{N,H,O\}\)
is defined as the average pairwise distance:
\begin{equation}
H^{(\ast)} = 
\frac{2}{N(N-1)} 
\sum_{1 \le i < j \le N} 
d\big(\phi_i^{(\ast)}, \phi_j^{(\ast)}\big),
\label{eq:pairwise_dis}
\end{equation}

where \(d(\cdot,\cdot)\) is a suitable distance metric.

The overall swarm heterogeneity is a weighted sum:
\begin{equation}
H_{\text{total}} 
= 
\alpha_N H^{(N)} 
+ 
\alpha_H H^{(H)} 
+ 
\alpha_O H^{(O)},
\label{eq:wt_sum}
\end{equation}

with non-negative weights 
\(\alpha_N, \alpha_H, \alpha_O \ge 0\).

\subsection{Task Assignment Optimization Formulation}

% --- Task Assignment Optimization Formulation ---
Let \(\mathcal{T} = \{1,2,\dots,M\}\) be a set of tasks and
let \(c_{ij}\) denote the cost for agent \(i\) to execute task \(j\).
Define binary assignment variables \(a_{ij} \in \{0,1\}\).

The task allocation problem is
\begin{equation}
\min_{\{a_{ij}\}}
\sum_{i=1}^{N} \sum_{j=1}^{M} c_{ij} a_{ij},
\label{eq:alloc}
\end{equation}

subject to the constraints
\begin{equation}
\sum_{i=1}^{N} a_{ij} = 1,
\quad \forall j \in \mathcal{T},
\label{eq:constraints}
\end{equation}

\[
a_{ij} = 0
\quad \text{if agent } i \text{ lacks the capability for task } j.
\]

In heterogeneous swarms, the feasibility of assignments and the cost coefficients \(c_{ij}\)
depend explicitly on 
\(\phi_i^{(N)}, \phi_i^{(H)}, \phi_i^{(O)}\).  Extensions of this formulation to accommodate asynchronous execution, characterized by heterogeneous task completion times across agent types, have been explored using macro action decomposition and distributed reinforcement learning~\cite{10301527}.

\subsection{Resilience and Performance Metrics}
% ---  ---
Let \(J\) be a mission performance metric for a swarm configuration \(\mathcal{S}\).
The nominal performance without failures is
\begin{equation}
J_{\text{nom}}(\mathcal{S}) 
= 
\mathbb{E}\big[\, \mathcal{P}(\mathcal{S}, \omega) \,\big],
\label{eq:mis_per}
\end{equation}
where \(\mathcal{P}(\mathcal{S}, \omega)\) is the mission outcome under scenario \(\omega\),
and the expectation is over environmental uncertainties.

Let \(\mathcal{F}\) be a set of failure modes.
For each \(f \in \mathcal{F}\), denote the performance under failure as \(J_f(\mathcal{S})\).
A resilience index is defined as
\begin{equation}
R(\mathcal{S})
=
\frac{1}{|\mathcal{F}|}
\sum_{f \in \mathcal{F}}
\frac{J_f(\mathcal{S})}{J_{\text{nom}}(\mathcal{S})}.
\label{eq:fail_modes}
\end{equation}

A heterogeneous swarm improves resilience when
\[
R(\mathcal{S}_{\text{hetero}})
>
R(\mathcal{S}_{\text{homo}}).
\]

To strengthen the theoretical foundation for heterogeneous swarm control, several developments in control techniques have been made. The coordination protocols triggered by events ~\cite{LI2020108898} coordinate network communication by invoking higher-level coordination for specific events. Such systems are beneficial in low-bandwidth heterogeneous systems. The methods of collaboration at a fixed time and at a prescribed time~\cite{9866828} guarantee convergence within a fixed time interval regardless of initial conditions. Guaranteed performance guarantees can be offered for missions with strict time requirements. The theoretical advancements discussed on coordination under constraints, at some convergence points (and failure events), complement the above task assignment and resilience frameworks by examining their evolution over time.

\section{Classification of Heterogeneous Swarms}
\label{Classify}
Heterogeneity can manifest in various forms, and researchers have employed the term ``heterogeneous swarms'' to describe a range of diverse swarm deployments. This study broadly classifies heterogeneous swarms into three major sections:

\begin{enumerate}
\item \textbf{Nature:} when agents differ from other agents in the swarm in terms of their behavior and function. The agents may have the same hardware, but each is assigned a different role.
\item \textbf{Structure:} when noticeable differences in structure exist between the swarm agents. For example, a formation of fixed-wing high altitude UAV (Uncrewed Aerial Vehicle) agents working in conjunction with low-level quadcopters.
\item \textbf{Operational space:} when there are differences in the operational space that the agents in a swarm will operate in. Multi-space hetero swarms may exist wherein agents in the air, on the ground, and on the water surface coordinate to accomplish a mission.
\end{enumerate}

\begin{figure*}[!t]
    \centering
    \includegraphics[width=0.9\textwidth]{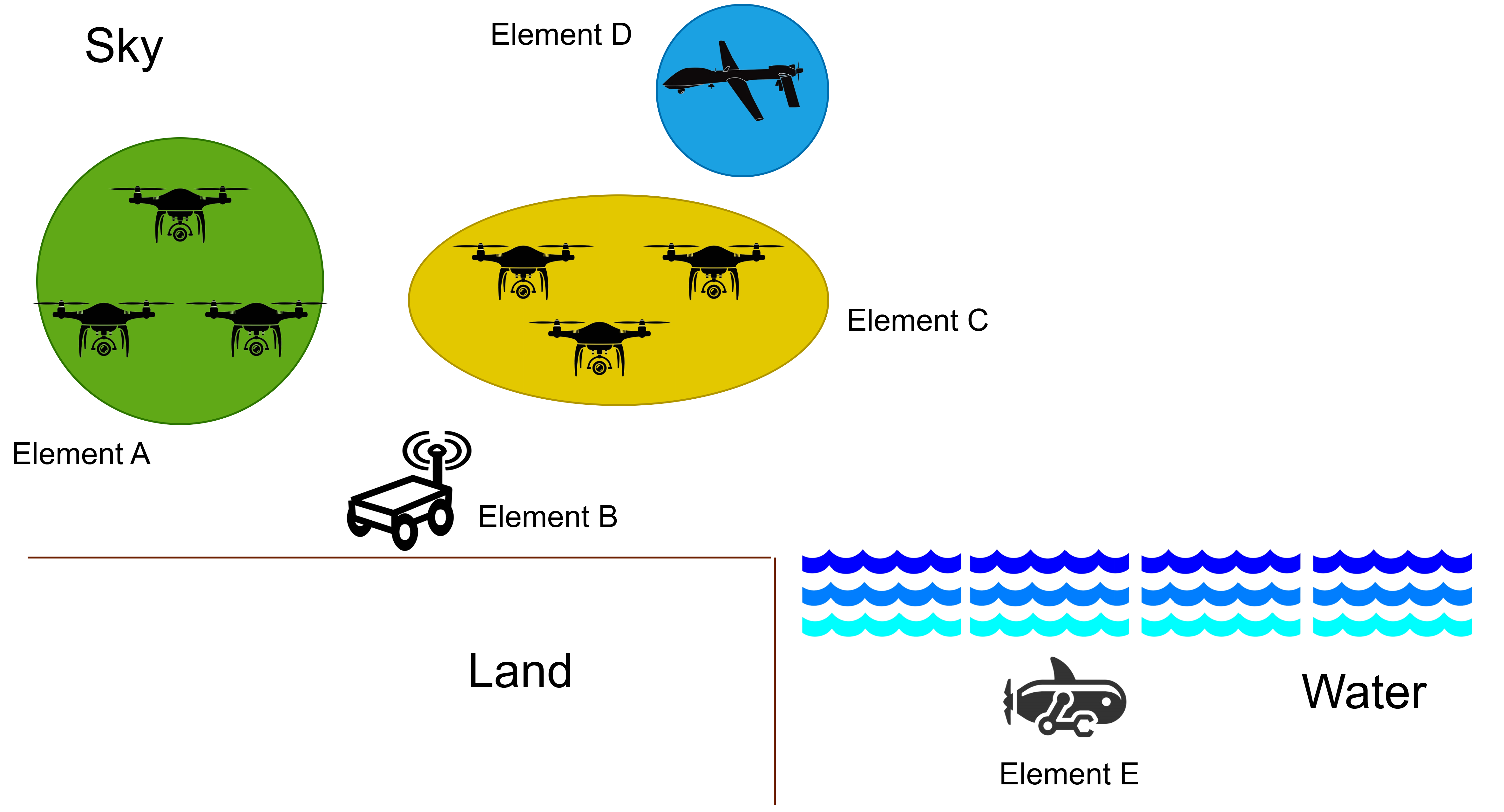}
    \caption{Diagram highlighting multiple swarm configuration possibilities}
    \label{fig:fig3}
\end{figure*}

Heterogeneous swarms are classified according to their nature, structure, and operational space. Figure \ref{fig:fig3} illustrates the various ways in which a heterogeneous swarm can be composed. The vehicles are described in different elements. Element A is a swarm of quadcopter UAVs that are similar in structure and functionality; i.e., they are a homogeneous swarm. Element B is an uncrewed ground vehicle (UGV). Element C is a group of quadcopters that are similar in structure to Element A but are assigned different roles. Element D is a fixed-wing high-altitude UAV. Element E is an uncrewed underwater submersible. Element C is a heterogeneous swarm in itself. For example, some agents in the swarm may serve as attack agents, while others are programmed to be defensive. Such differences in agents may not be visible in their structure but rather in their behavior during swarm operations. Elements A and D form a heterogeneous single-space swarm, whereas if elements B, E, or both are also added, they result in a multi-space heterogeneous swarm. Multiple such combinations can be used to create heterogeneous swarms.

The three categories described are nature, structure, and operational space, which often interact in practical swarm deployments. Real-world heterogeneous swarms frequently include overlapping types of diversity. For example, agents may share the same operational space but differ in both hardware and behavioral roles, or they may differ in operational space and also possess distinct control architectures.

Recognizing this overlap is important for researchers and designers because it influences:
\begin{itemize}
    \item Communication architecture: heterogeneous teams often rely on multi-layer communication models to bridge different sensing frequencies, bandwidth availability, and latency constraints.

    \item Task allocation strategies: the type of heterogeneity directly shapes how tasks can be distributed. For instance, role-based heterogeneity allows dynamic reallocation, while hardware-based heterogeneity may enforce fixed responsibilities.

    \item Resilience assessment: different types of heterogeneity contribute to resilience in various ways. Operational-space heterogeneity provides multi-angle coverage and redundancy, whereas behavioral heterogeneity improves adaptability.

    Recent research addresses the interdependencies through unified frameworks: graph neural network methods~\cite{goarin2024graph} facilitate communication-efficient task allocation under bandwidth constraints, while multi-modal SLAM systems~\cite{chang2022lamp} ensure robust localization across diverse sensing modalities in GPS-restricted environments, illustrating that systematic integration of heterogeneous types can minimize individual limitations.

\end{itemize}

\subsection{Identification Based on Agent Nature}

Agents in a swarm that all operate in the same operational space but have different behaviors and functions fall into this category. Examples include pursuer-evader agents in game-theoretic approaches inspired by observations of natural phenomena, such as predators hunting or flocks of birds in flight. Such agents are heterogeneous in terms of capabilities rather than hardware. Although agent hardware can affect capabilities such as speed, flight altitude, and offensive capacity, this section examines how homogeneous agents are converted to heterogeneous ones. There is a large body of research on agents that are structurally similar but modeled with different constraints or behavior functions, thereby making them unique within their swarm. The authors in \cite{Makkapati2020ApolloniusAA} assigned the most apparent characteristic to swarm agents by giving them different speed capabilities. To solve the multi-agent pursuit problem, the pursuers employ a dynamic task-allocation algorithm to capture an evader agent that can employ any necessary strategy to evade capture. Studies on homogeneous agents trained with graph neural networks and reinforcement learning, such as those in \cite{deka2020naturalemergenceheterogeneousstrategies}, depict the natural emergence of heterogeneous behavior that is entirely independent of the agents' hardware. The system proposed in \cite{deng2013cooperative} presents a mission-planning model for heterogeneous UAVs that uses multi-objective particle swarm optimization (MOPSO). Study \cite{deng2013cooperative} investigated the task assignment problem for UAVs with varying capabilities and constraints, utilizing a modified genetic algorithm to capture target features for search UAVs effectively. This study used a set of fixed-wing UAVs with limited resources. Their heterogenetic identification is based on the UAV's available resources. UAV characteristics were well-defined by \cite{1501630}, where each UAV can have a personality of ``social searcher, antisocial searcher, or relay.'' The authors in \cite{1501630} assigned different tolerances to collision and avoidance beacons for agents in the swarm, using a single metric, alpha. Variations in these metrics result in different nature agents, even though they are structurally similar. Such a multi-agent swarm is then used in detecting chemical clouds. The nature of agents is defined differently across these studies; however, introducing heterogeneity in agents' nature has positive effects on resilience in swarm operations.

Nature-based heterogeneity is ubiquitous in swarms where hardware uniformity is preferred for cost, ease of replacement, or logistical reasons. Several additional forms of nature-based heterogeneity include:

\begin{itemize}
    \item Priority-based behavior: agents can be assigned mission priorities, such as “first responder,” “support,” or “reserve,” which influence the order in which they engage tasks.
    \item Adaptive aggressiveness: some agents take high-risk paths (e.g., for faster target pursuit), while others adopt more conservative strategies to preserve swarm stability.
    \item Information-driven roles: agents may specialize as local mappers, global trackers, anomaly detectors, or communication relays.
    These behaviors may be implemented through rule-based controllers, optimization algorithms, or learned policies.
    \item Learning-driven coordination: Prior work, such as \cite{9508420}, shows that developing coordination strategies through reinforcement learning based on teammate behavior and environmental dynamics yields learned interactions that enable homogeneous hardware to exhibit heterogeneous coordination patterns.
    
\end{itemize}

This behavioral heterogeneity strengthens resilience by enabling the swarm to maintain mission progress even when specific roles fail, adapt strategies without requiring hardware changes, and reassign roles dynamically based on agent health or task urgency. Recent progress in learning based methods shows that training can naturally create behavioral roles~\cite{10301527}. For example, agents with the same hardware can learn to collaborate in novel ways through asynchronous multi-agent reinforcement learning, thereby eliminating the need for predefined role assignments.

\subsection{Identification Based on Agent Hardware}

\begin{figure*}
    \centering
    \includegraphics[width=1\linewidth]{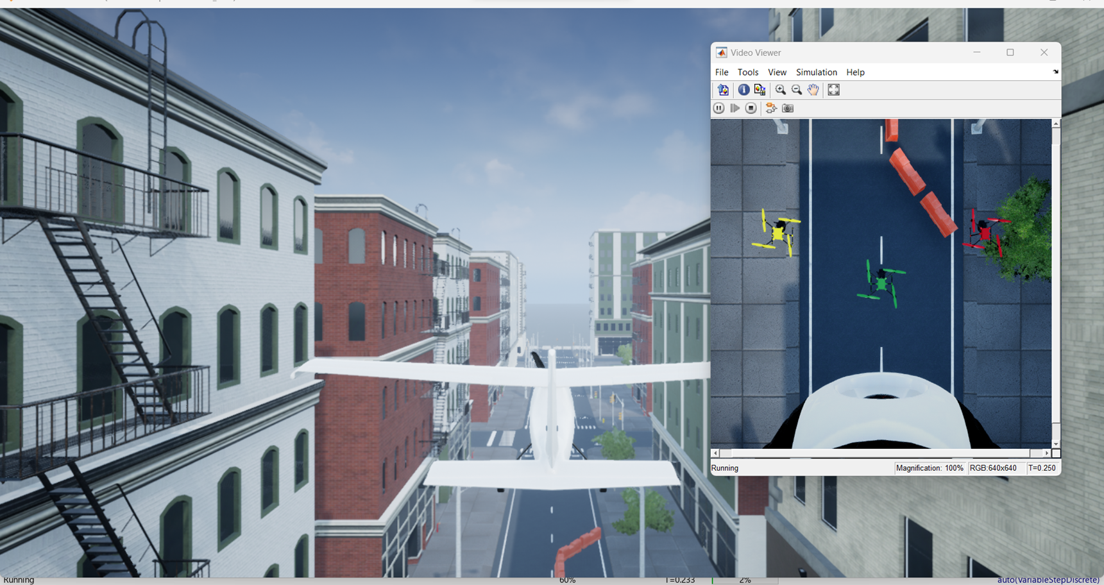}
    \caption{Simulation instance of quadcopter UAV swarm followed by a fixed-wing UAV}
    \label{fig:matlab}
\end{figure*}

Labeling heteroswarms by hardware and operating space often overlaps, since different hardware is typically required for agents to operate in different operational spaces. There are instances in which swarms comprise other types of agents in the same operational space, such as a flock of quadcopters working in tandem with a high-altitude fixed-wing aircraft that handles communication and swarm management [as shown in Fig. \ref{fig:matlab}]. However, fewer additional examples of similar swarms operating in air or water are reported in the existing literature, and those that are reported exhibit different capabilities. Swarms of water-cleaning vehicles are in development, with some agents performing cleanup duties and others maintaining swarm formation, detecting the spread of water pollutants, and recovering damaged cleanup vehicles, among other tasks. The authors of \cite{7271613} described sea-based exercise missions in which uncrewed aircraft systems (UAS) collaborate with uncrewed underwater vehicles (UUV). This is an example of a multi-space hetero swarm with control nodes on the shore that support geospatial data and mission planning. The authors claimed that this was the first successful demonstration of messaging protocols in cross-domain uncrewed systems. Several teams, such as \cite{7849525, 10.1007/s10846-009-9334-x, app10051583}, demonstrated excellent coordination between UWSVs (Uncrewed Water Surface Vehicles) and UAV swarms, in which the surface vehicles served as mobile base platforms for the UAVs, thereby extending the UAVs' range at sea. Unlike traditional UAV swarms that may need to return to shore to recharge and redeploy, these UWSV platforms can provide fuel support to the UAV at sea itself, thereby eliminating the need to return to shore every time.

Multi-level swarms were described in \cite{7849525}, in which heterogeneous aircraft were used for surveillance. Such swarms had agents operating at different levels within the operational space, or in entirely different operational spaces. The authors in \cite{10.1007/s10846-009-9334-x} described a comprehensive command-and-control architecture for a fleet of micro-aircraft, supported by higher-altitude, endurance-capable aircraft called the ``Ares'' aircraft. Both types of agents differ in structure, size, and flight capabilities, resulting from their hardware.

Hardware heterogeneity often arises from deliberate design decisions that align each agent platform with a specialized role. For example:

\begin{itemize}
    \item Fixed-wing UAVs provide wide-area coverage and persistent surveillance.
    \item Quadcopters provide high-precision maneuvering, hovering, and close-range sensing.
    \item UGVs carry larger batteries, can transport equipment, and can offload computational tasks.
    \item UWSVs and UUVs contribute environmental sensing capabilities that aerial or ground agents lack, such as sonar or water-quality sampling.
\end{itemize}

This variety enables the swarm to refine its operational granularity and combine macro-level awareness (e.g., high-altitude imaging) with micro-level analysis (e.g., ground or underwater inspection). Figure \ref{fig:matlab} shows a MATLAB simulation from prior work by the authors, where 3 low-altitude quadcopters are surveying an urban area. A fixed-wing UAV flying above the buildings tracks the UAVs and manages communication. Hardware heterogeneity also enables the use of specialized sensors, which are essential for reliable perception under adverse conditions. For example, combining ground robots with LiDAR (Light Detection and Ranging) and aerial platforms with vision can help them navigate underground areas where GPS signals are weak~\cite{chang2022lamp,tranzatto2022cerberusautonomousleggedaerial}. This is because no single sensor type can reliably perform mapping and navigation on its own.

\subsection{Identification Based on Operational Space}

Operational space is the target area in which a UV swarm executes its mission. Uncrewed vehicles have been used to perform functions on land, water, underwater, and in the air. The coordination of surface vehicles has found widespread applications. Proposed ideas include the creation of UGV-UAV systems. An uncrewed surface vehicle (USV) is a term commonly used to describe vehicles that operate on both land and water \cite{app10051583}. For example, a ground rover and an uncrewed boat vessel both fall under the broad USV category. Single-space hetero swarms operate in a single operational space (i.e., air, water, or ground). Such swarms are solely comprised of UAVs, UGVs, or UWSVs. Although individual agents in the swarm may have distinct characteristics or hardware traits that differentiate them from one another, their operational space remains the same. Such swarms are typically used for focused applications where venturing into other areas is not necessary. Some applications require deploying agents across multiple spaces, such as air-water, air-ground, or surface-water \cite{6107312}. By their very nature, swarms operating across multiple operational spaces will be heterogeneous, as agents require different behaviors and hardware to function in their diverse environments, such as air, land, surface, underground~\cite{6107312}, water, or underwater. Swarms operating across multiple operational spaces will differ in nature because agents require distinct behaviors and hardware to operate in different environments, such as air, land, surface, underground, water, or underwater. Multi-space heterogeneity has been especially useful in precision agriculture, where aerial drones monitor crops and capture multispectral images. At the same time, ground rovers perform targeted tasks such as precision spraying and soil sampling~\cite{JU2022107336}. Coordination across domains enables resource optimization and a reduced environmental impact. As demonstrated by the authors of this study in a currently underway experiment, UGVs have shown significant success in replacing human observers in E-VLOS (Extended Visual Line of Sight) flights conducted by UAVs, thereby reducing reliance on humans \cite{Phadke_2026}.

\subsection{Classifying Current Research}

Heterogeneous agent research mentioned throughout this review is summarized in Table \ref{tab:classification}. Using the definitions proposed in this study, the studies are identified as heterogeneous swarms either by nature and/or by hardware. The studies are also classified according to whether the swarms operate in single or multiple spaces. During classification, agents operating at different levels in the same space are identified as heterogeneous by hardware, not by nature. This is due to the fact that operating at multiple levels may require several modifications to the agent's structure, without which its function would be impossible. For example, a UAV operating at a higher altitude in a swarm can be a fixed-wing aircraft, whereas lower-level drones are quadcopters. This difference in altitude is thus due to the fixed-wing's higher flight ceiling than the quadcopter's. Similarly, a water-surface vehicle might not be able to submerge due to the lack of hardware that a submersible has. However, a submersible might be able to act as a surface vehicle for some time.

A notable limitation across many studies summarized in Table \ref{tab:classification} is the lack of standardized performance metrics for evaluating heterogeneous swarms. While some works report improvements in mission completion time, detection probability, or coverage area, others rely on qualitative assessments or scenario-specific success criteria. This diversity of evaluation methods complicates direct comparison between approaches and highlights the need for unified benchmarking frameworks tailored to heterogeneous multi-agent systems. The DARPA Subterranean Challenge~\cite{tranzatto2022cerberusautonomousleggedaerial} developed scenarios for the competition-based evaluation of team performance under realistic failure conditions. Benchmarks based on learning evaluate the quality of coordination via graph metrics ~\cite{goarin2024graph}.

In addition to quantitative benchmarking on selected competitions, recent work systematically examines the extent of heterogeneity in swarm applications. Comprehensive reviews related to swarm intelligence techniques~\cite{appliedmath4040064} and artificial intelligence-enabled task assignment methods~\cite{Li2024} show recurring tendencies towards learning-based coordination and adaptive team formation. Research on foraging efficiency~\cite{10.3389/frobt.2024.1426282} reveals that heterogeneous teams can outperform homogeneous ones. They also present a theoretical study of an actual swarm robotics mission. Domain-specific studies on disaster response systems~\cite{10.3389/frobt.2024.1362294} and global exploration. It follows from these studies that heterogeneity is moving from a matter of curiosity to a standard design consideration across a range of robotic applications.

\begin{table*}[!t]
\caption{Classification of Heterogeneous Swarm Research}
\label{tab:classification}
\centering
\footnotesize
\renewcommand{\arraystretch}{1.2}
\begin{tabular}{|c|p{8.5cm}|c|c|c|c|}
\hline
\textbf{Ref} & \textbf{Research Contribution} & \textbf{Nature} & \textbf{Hardware} & \textbf{Single} & \textbf{Multi} \\
\hline
\cite{López17} & Mission planning and monitoring for hetero swarms & $\checkmark$ & $\checkmark$ & $\checkmark$ & $\checkmark$ \\
\hline
\cite{6731679} & Swarming of hetero swarms for SAR in hostile territory & $\checkmark$ & $\checkmark$ & $\checkmark$ & --- \\
\hline
\cite{1501630} & Hetero swarms are used in the detection of chemical clouds & $\checkmark$ & --- & $\checkmark$ & --- \\
\hline
\cite{9215994} & Swarming behavior of hetero vehicles for SAR & $\checkmark$ & --- & $\checkmark$ & --- \\
\hline
\cite{7761074} & Multi-space hetero swarm communication and collaboration & --- & $\checkmark$ & --- & $\checkmark$ \\
\hline
\cite{app10051583} & UAV landing and recovery using UWSV & --- & $\checkmark$ & --- & $\checkmark$ \\
\hline
\cite{7271492} & Interoperability in heterogeneous swarm vehicles & --- & $\checkmark$ & --- & $\checkmark$ \\
\hline
\cite{deng2013cooperative} & Task assignment for gene-modified hetero swarms & $\checkmark$ & --- & $\checkmark$ & --- \\
\hline
\cite{s20185026} & Task planning for hetero UAV vehicles & $\checkmark$ & --- & $\checkmark$ & --- \\
\hline
\cite{app11199145} & Cross-space formation control and containment & --- & $\checkmark$ & --- & $\checkmark$ \\
\hline
\cite{10.3389/frobt.2021.616950} & Multi-space vehicle surveillance operations & --- & $\checkmark$ & --- & $\checkmark$ \\
\hline
\cite{8491531} & Consensus formation control & --- & $\checkmark$ & $\checkmark$ & --- \\
\hline
\cite{deka2020naturalemergenceheterogeneousstrategies} & Emerging capabilities in hetero swarms & $\checkmark$ & --- & $\checkmark$ & --- \\
\hline
\cite{8995977} & Distributed task allocation & $\checkmark$ & --- & $\checkmark$ & --- \\
\hline
\cite{8833293} & Mission planning using improved multi-objective PSO & $\checkmark$ & --- & $\checkmark$ & --- \\
\hline
\cite{jmse9111314} & Control scheme for hetero swarms & --- & $\checkmark$ & --- & $\checkmark$ \\
\hline
\cite{9389145} & Target search and track & $\checkmark$ & --- & $\checkmark$ & --- \\
\hline
\cite{9256688} & Optimal control techniques & $\checkmark$ & $\checkmark$ & $\checkmark$ & --- \\
\hline
\cite{10.5555/2484920.2485103} & Mapping application using multi-space vehicles & --- & $\checkmark$ & --- & $\checkmark$ \\
\hline
\cite{10.1007/s10846-009-9334-x} & Communication and control protocols & --- & $\checkmark$ & $\checkmark$ & --- \\
\hline
\cite{7849525} & Multi-agent- multi-level hetero swarms for surveillance & --- & $\checkmark$ & $\checkmark$ & --- \\
\hline
\cite{Makkapati2020ApolloniusAA} & Evader-pursuer problem & $\checkmark$ & --- & $\checkmark$ & --- \\
\hline
\cite{rs12101608} & Terrain mapping using heterogeneous vehicle teams & --- & $\checkmark$ & --- & $\checkmark$ \\
\hline
\cite{9264359} & Formation tracking for hetero swarms & --- & $\checkmark$ & --- & $\checkmark$ \\
\hline
\cite{9237985} & Combat mission planning using swarm optimization for hetero teams & --- & $\checkmark$ & --- & $\checkmark$ \\
\hline
\cite{chang2022lamp} & Multi-robot SLAM for underground environments & --- & $\checkmark$ & $\checkmark$ & --- \\
\hline
\cite{JU2022107336} & Multi-robot systems in agriculture review & $\checkmark$ & $\checkmark$ & --- & $\checkmark$ \\
\hline
\cite{tranzatto2022cerberusautonomousleggedaerial} & CERBERUS: Legged and aerial exploration (DARPA SubT) & --- & $\checkmark$ & --- & $\checkmark$ \\
\hline
\cite{goarin2024graph} & Graph neural network for decentralized goal assignment & $\checkmark$ & --- & $\checkmark$ & --- \\
\hline
\cite{10301527} & Asynchronous multi-agent reinforcement learning & $\checkmark$ & $\checkmark$ & $\checkmark$ & --- \\
\hline
\cite{horyna2023decentralized} & Decentralized UAV swarms for SAR without communication & $\checkmark$ & --- & $\checkmark$ & --- \\
\hline
\cite{deng2023distributed} & Distributed reconnaissance and strike task allocation & $\checkmark$ & --- & $\checkmark$ & --- \\
\hline
\cite{9508420} & Neural-Swarm2: Planning with learned interactions & $\checkmark$ & $\checkmark$ & $\checkmark$ & --- \\
\hline
\cite{WANG2024315} & Air-ground coordinated unmanned swarm systems & --- & $\checkmark$ & --- & $\checkmark$ \\
\hline
\cite{ai5020029} & Attention mechanism for task-adaptive robot teaming & $\checkmark$ & $\checkmark$ & $\checkmark$ & --- \\
\hline
\cite{BARTH2024689} & Planetary exploration using deep reinforcement learning & $\checkmark$ & $\checkmark$ & --- & $\checkmark$ \\
\hline
\end{tabular}
\end{table*}

The classification framework outlined in this section, comprising nature, hardware structure, and operational space, offers a systematic approach to analyzing heterogeneous swarm designs. To understand how these types of heterogeneity show up in real systems, specific implementations need to be examined. The next section examines deployments across different application areas, showing how researchers leverage and combine these heterogeneities to meet the needs of specific missions.

\section{Effects of heterogeneity on Operational factors}
\label{op}
\subsection{Implementation Considerations}

The operation of a heterogeneous swarm comes with its own complexities. Controlling a single space homogeneous swarm is complicated enough. The authors' finding in \cite{8833293} is that most current research on mission planning and swarm dynamics focuses on homogeneous UAVs. Hence, existing algorithms such as particle swarm optimization (PSO), game-theoretic operation techniques, or swarm management tools cannot be applied. When considering multi-space swarms, attention should be given to all dynamic operational spaces. For homogeneous swarms, this is usually not required. While a UGV swarm does not require tide and current data, a UGV-UWSV needs to take into consideration both shoreline terrain data for the UGV, as well as tide data for the UWSV. The addition of diverse agents introduces additional factors, including the dynamics of the new operational space, new communication protocols, and diverse capabilities. Interoperability issues are addressed in projects such as ICARUS \cite{7761074}. Marsupial platforms for UAV-UWSVs, such as those in \cite{7761074, 7271492}, must address key problems, including task management and UAV-agent recovery.

The solutions developed in \cite{app10051583} propose a fully autonomous recovery system for fixed-wing aircraft landing on ships, using arresting cables and nets to recover aerial agents. Such techniques can effectively capture multiple agents, thereby enabling the deployment of numerous aircraft and their recovery from a single platform. The authors of \cite{app10051583} noted that coordinating the arresting cable system on the recovery boat with the aircraft's UAV is challenging. Existing time-varying formation-tracking methods have been demonstrated only for homogeneous systems, as in \cite{8700227}. The authors of \cite{9264359} cited the lack of research on formation control as a motivation for developing improved formation-tracking algorithms for multi-space systems. A general trend observed is that essential functions, such as consensus, mission planning, resource optimization, and task assignment, are present only in homogeneous systems. Hence, the different terrain characteristics need to be taken into account, along with the development of ground-up policies for heterogeneous systems operating across various terrains. In formation tracking, predefined formations often do not serve the purpose. Switching to a different topology \cite{9264359} also requires consideration of communication protocols.

The inclusion of heterogeneous agents significantly increases the complexity of swarm control and coordination. Agents operating in different domains or using distinct kinematic models require specialized controllers that must be harmonized at the swarm level. For instance, control laws suitable for aerial vehicles may be unsuitable for ground or underwater vehicles due to differences in dynamics, actuation limits, and environmental disturbances. To address this, many heterogeneous swarm systems adopt hierarchical or modular control architectures, in which low-level controllers are tailored to individual platforms, while high-level planners coordinate inter-agent behavior. This separation of concerns enables scalable control while allowing each agent to operate optimally within its domain.

Communication remains a critical challenge in heterogeneous swarms, particularly in multi-space deployments. Differences in communication modes, such as RF links for aerial vehicles and acoustic links for underwater vehicles, introduce latency, bandwidth, and reliability issues. These differences often necessitate the use of relay agents or gateway nodes that translate between communication protocols and maintain global situational awareness. Furthermore, heterogeneous agents may have asymmetric communication capabilities, with some agents able to broadcast widely while others operate with limited connectivity. Designing robust communication strategies that accommodate these asymmetries is critical for maintaining coordination, especially in contested or cluttered environments. Recent work, such as ~\cite{chang2022lamp,tranzatto2022cerberusautonomousleggedaerial}, has shown that integrating LiDAR, visual, and inertial data from different platforms can mitigate communication failures by enabling local autonomous decision-making and map sharing during short periods of connectivity.

Energy constraints play a decisive role in heterogeneous swarm operations. Agents with high mobility or heavy payloads often consume energy at significantly different rates, affecting mission endurance and scheduling. Heterogeneous swarms must therefore incorporate energy-aware planning strategies that account for varying battery capacities, recharging requirements, and refueling opportunities. In multi-space systems, surface or ground vehicles frequently serve as mobile support units, providing recharging or maintenance capabilities to aerial agents. This introduces additional planning layers related to rendezvous timing, docking mechanisms, and task handoffs.
In agricultural heterogeneous systems, ground rovers with higher battery capacity serve as mobile charging stations for aerial drones conducting crop surveys, enabling energy-aware coordination~\cite{JU2022107336}. Maintaining consistent aerial-to-ground communication based on battery life estimates and task assignments prevents mission termination and enables longer missions.

When mixed groups operate across large geographic areas, network connectivity and routing face additional challenges. Adaptive performance and understanding of the diversity in agents' communication capabilities and mobility patterns are crucial. AI-based routing protocols have been developed by ~\cite{ROVIRASUGRANES2022102790}. For instance, fixed-wing UAVs flying at high altitudes may be able to communicate over long distances. However, rotorcraft operating at lower altitudes can communicate over limited distances. When aerial agents have limited flight time, energy-aware routing is crucial. New optimization approaches ~\cite{10414185} show that a UAV's speed can be adjusted based on the required coverage area, thereby extending the mission duration. Approaches for ambiently powered robot swarms ~\cite{MOKHTARI2025104898} have shown that intelligent task scheduling can better utilize a range of energy-harvesting capabilities, including solar-powered ground stations with ground batteries and aerial scouts.

Localization and mapping features in heterogeneous deployments need special attention. Active SLAM methods~\cite{10.1109/TRO.2023.3248510} allow robots to choose smart sensing actions to gather more information about the environment to increase map gain. Having different types of agents involved can be particularly useful, as it provides diverse perspectives. For instance, aerial vehicles can be deployed to monitor a large area and quickly identify sites of interest, while ground vehicles can be used for closer inspection of specific sites.  To ensure the proper functioning of different sensor types, planning is required to balance exploration speed and map accuracy. Active SLAM methods use a decision-making approach for localization. Agents consider which moves to make at the optimal time to maximize utility. They consider how much information they will get in the future. This is done using sensing actions and sensing movement-path choices.

\subsection{Increased Resilience in Heterogeneous Swarms}

The following section reviews research showing that introducing heterogeneity increases swarm performance. This can be single or multi-space, or by nature or hardware capability. Various swarm characteristics improved with the inclusion of hetero agents. These characteristics are attributed to the inclusion of specific agents that exhibit increased movement speed, possess evolved capabilities, enhance task performance, and leverage complementary strengths within the swarm.

Resilience in heterogeneous swarms extends beyond simple fault tolerance. It is the swarm's ability to adapt its structure, roles, and coordination strategies in response to internal failures or external disturbances. Heterogeneity enables and enforces resilience by distributing critical functions across agents with distinct capabilities, reducing the likelihood that a single point of failure will compromise the entire mission. From a systems perspective, resilience can be viewed as an emergent property resulting from redundancy, diversity, and adaptability. Behavioral heterogeneity allows rapid role reassignment, hardware heterogeneity provides functional redundancy across platforms, and operational-space heterogeneity enables spatial diversification of sensing and actuation.

Heterogeneous swarms are particularly effective at mitigating common failure modes, including agent loss, sensor degradation, and communication breakdowns. For example, if an aerial scout fails during a mission, a ground-based agent equipped with alternative sensing modalities may partially compensate by continuing localized exploration. Similarly, communication disruptions can be mitigated by agents that act as mobile relays or data ferries. Recovery strategies in heterogeneous swarms often involve dynamic reconfiguration, in which remaining agents adjust their roles, formations, or task assignments to preserve mission objectives. These adaptive behaviors are more difficult to achieve in homogeneous systems, where all agents share identical limitations.

\subsubsection{Increase in Movement Speed}
Several studies, such as \cite{Makkapati2020ApolloniusAA}, have assumed that pursuers exhibit faster motion than evaders. Indeed, in most agent scenarios, achieving the required pursuit speed requires lightweight agents with only the most basic onboard equipment. However, a military multi-functional swarm application would require agents equipped with additional communication and defense equipment. Here, a swarm comprising quicker pursuer vehicles backed by heavier defensive vehicles is effective. Similarly, experiments conducted by \cite{9389145} on heterogeneous swarms involve agents with varying speeds and employ velocity functions to assess the swarms' responses to the environment. The results showed that swarm responsiveness gradually increased with the proportion of faster agents in the swarm.
In heterogeneous swarms, differences in agent speed can be strategically exploited to balance rapid response with sustained operation. Faster agents may be deployed for initial reconnaissance or pursuit tasks, while slower but more capable agents provide persistent coverage or logistical support. This separation allows the swarm to react quickly to emerging threats without exhausting critical resources.

Performance improvements can be observed even in low-level experiments \cite{phadke2023designing}.

\begin{figure*}
    \centering
    \includegraphics[width=1\linewidth]{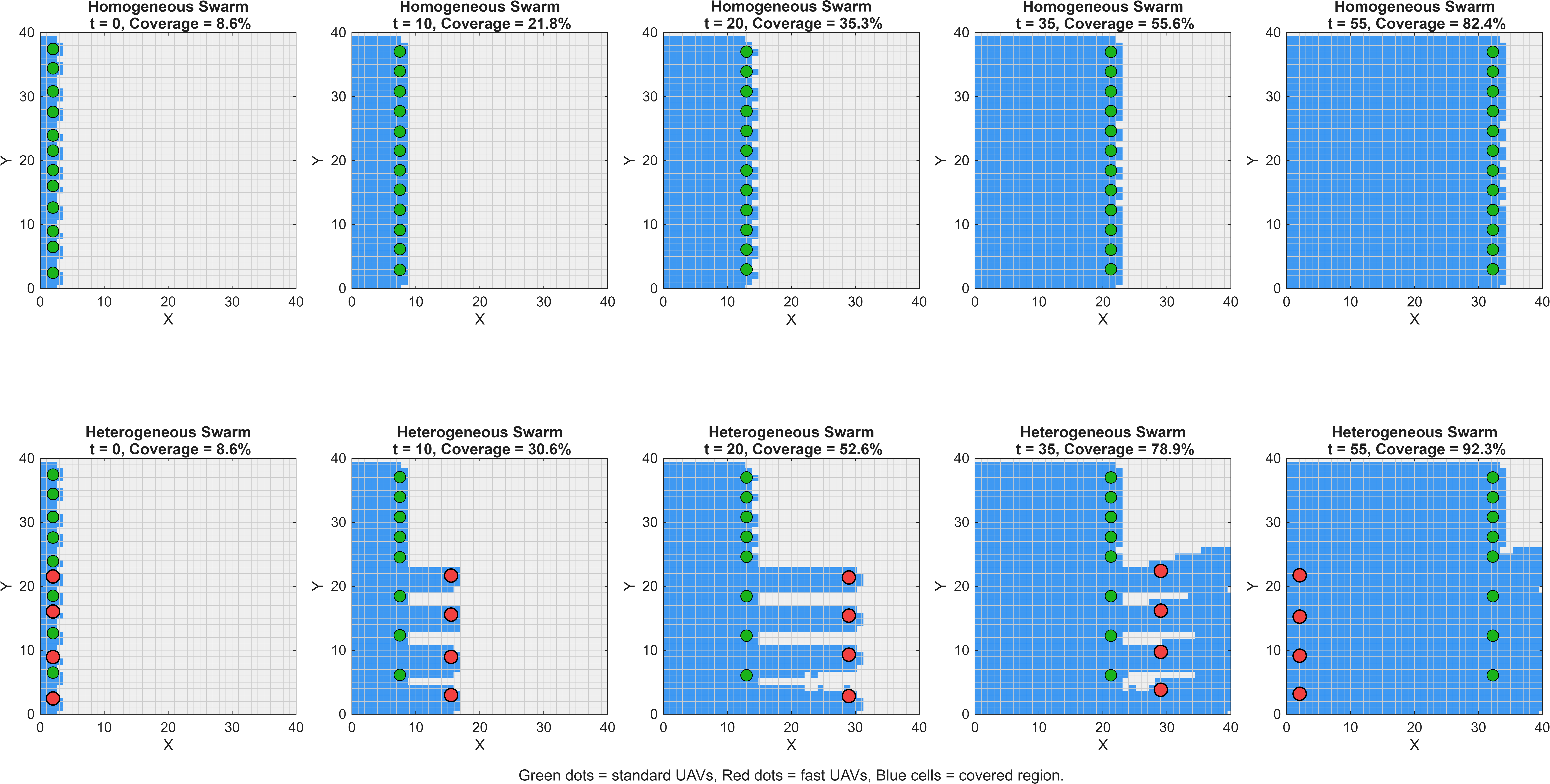}
    \caption{Static time-series comparison of area coverage for homogeneous and heterogeneous UAV swarms over a 2D grid map. The top row shows a homogeneous swarm in which all UAVs move with the same speed. The bottom row shows a heterogeneous swarm in which at least 30\% of the UAVs are fast agents.}
    \label{fig:coverage_subplots}
\end{figure*}

\begin{figure}
    \centering
    \includegraphics[width=1\linewidth]{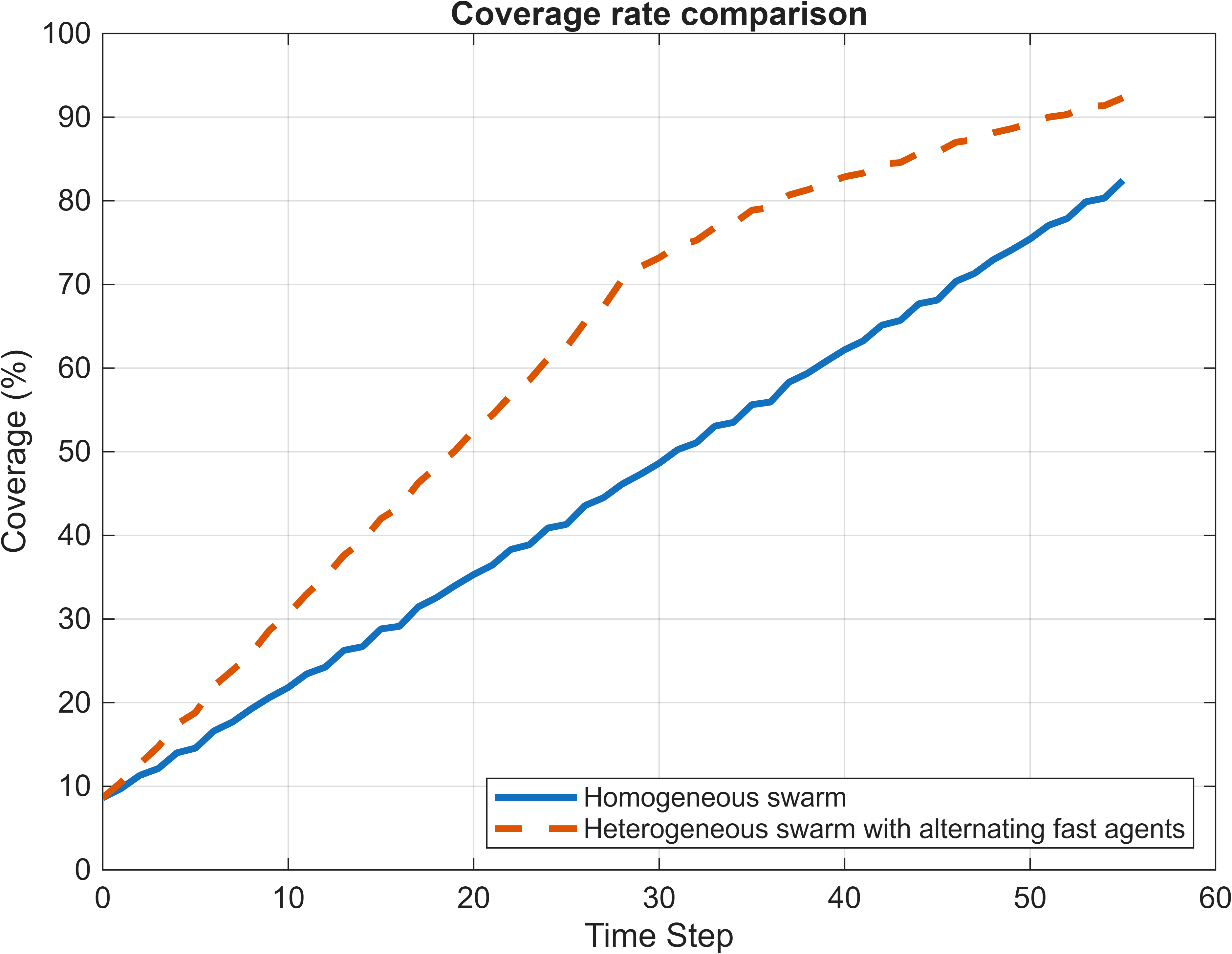}
    \caption{Coverage ratio as a function of time for the homogeneous and heterogeneous UAV swarms. The heterogeneous swarm achieves a faster rise in coverage because fast agents traverse unexplored regions more quickly, reducing the time required to observe the full grid.}
    \label{fig:coverage_curves}
\end{figure}

\begin{equation}
\mathcal{C}(t)=\bigcup_{i=1}^{N}\left\{ q \in \mathcal{Q} \; \middle| \; \|\mathbf{q}-\mathbf{p}_i(t)\| \le r_i \right\},
\label{eq:covered_set}
\end{equation}

\begin{equation}
\eta(t)=\frac{|\mathcal{C}(t)|}{|\mathcal{Q}|}, 
\qquad
\mathbf{p}_i(t+1)=\mathbf{p}_i(t)+\mathbf{v}_i(t)\Delta t,
\label{eq:coverage_ratio}
\end{equation}

Let $\mathcal{Q}$ denote the set of discretized cells in the 2D mission map, and let $\mathbf{p}_i(t)$ denote the position of UAV $i$ at time step $t$. A grid cell $q \in \mathcal{Q}$ is considered covered if it lies within the sensing footprint of at least one UAV. Accordingly, the covered region at time $t$ is defined by \eqref{eq:covered_set}, where $r_i$ is the sensing radius of agent $i$.

The overall coverage performance is denoted by the coverage ratio $\eta(t)$ in \eqref{eq:coverage_ratio}, which measures the fraction of the environment that has been observed up to time $t$. The UAV motion is modeled through a discrete-time kinematic update, where $\mathbf{v}_i(t)$ is the velocity of agent $i$ and $\Delta t$ is the time step. In the heterogeneous swarm, a subset of agents is assigned a higher speed, i.e., $\|\mathbf{v}_i(t)\| = v_f > v_s$, allowing the team to traverse previously uncovered regions more rapidly and thereby increase the growth rate of $\eta(t)$.

\begin{equation}
v_i=
\begin{cases}
v_f, & i \in \mathcal{F},\\
v_s, & i \in \mathcal{S},
\end{cases}
\qquad \text{with } v_f > v_s,
\label{eq:hetero_speed}
\end{equation}

Here, $\mathcal{F}$ and $\mathcal{S}$ denote the sets of fast and standard UAVs, respectively. The condition $v_f > v_s$ captures the heterogeneous mobility capability that enables faster global coverage.

Figure~\ref{fig:coverage_subplots} illustrates the spatial evolution of coverage for both swarm configurations. In the homogeneous case, the covered region grows steadily as all UAVs sweep the environment at the same rate. In contrast, the heterogeneous swarm exhibits visibly faster spatial expansion because a subset of fast agents reaches distant, uncovered cells earlier in the mission. This is further confirmed in Figure~\ref{fig:coverage_curves}, where the heterogeneous swarm yields a steeper coverage trajectory and attains a higher covered fraction at identical time steps. These results support the hypothesis that speed heterogeneity can improve exploration efficiency in hetero-swarm coverage tasks.

\subsubsection{Evolving Capabilities of Homogeneous Agents}
Conclusive research, such as \cite{deka2020naturalemergenceheterogeneousstrategies}, has shown that homogeneous systems can be converted into heterogeneous systems solely through the influence of characteristic factors. The inferences from such research indicate that heterogeneous systems, labeled solely by their agent nature, are pretty successful at formulating deception strategies to counter attacks. Attack resilience against external agents, particularly enemy drone systems, is essential for resilient swarm dynamics. An added advantage of such systems, as highlighted in \cite{deka2020naturalemergenceheterogeneousstrategies}, is that agent replacement does not require specialized agents, differentiated by hardware, to be on standby. For example, if a fast agent is damaged, it must be replaced with another fast agent to maintain the swarm's full capabilities. However, in scenarios where all agents use the same hardware, any new agent can be deployed as a replacement, with its role designated by the operational software.
The emergence of functional diversity via learning-based methods demonstrates that heterogeneity need not entail permanent differentiation. Temporarily induced behavioral heterogeneity enables swarms to explore diverse strategies while retaining the ability to revert to baseline behavior. This flexibility is particularly valuable in environments where mission requirements evolve unpredictably. Recent progress in asynchronous multi-agent reinforcement learning~\cite{10301527} shows that learned behavioral specialization can work with timing heterogeneity. For example, agents with the same hardware can create different macro-action policies tailored to their execution speeds, which reduces idle time by 8–16\% compared to synchronized coordination strategies.
Communication-aware approaches to reinforcement learning studies ~\cite{10107729} further show that swarms can be trained on the jamming-resistant coordination strategies. When jamming is simulated in swarm communication channels, the performance with learned policies shows a 40\% higher success rate against communication attacks than that with traditional disturbance-based communication protocols.

\subsubsection{Complementary Capabilities}
The addition of agents that differ from the others in the swarm has enabled their capabilities to complement each other. A swarm may have inherent weaknesses arising from its constituent agents. These issues can vary, such as a lack of specific communication protocols or a reduced range. Such weakness can be mitigated by employing hetero agents within the swarm. For example, insufficient communication protocols may be solved by using relay drones that connect ground stations to other swarm agents. The hardware of these relay drones is specifically chosen to support a wide range of communication possibilities. If UAV agents in a swarm have an insufficient operational range, this can be addressed by adding marine or ground vehicles that serve as landing and charging agents for the aerial vehicles. Using an array of agents with diverse kinematic capabilities is beneficial because they offer complementary advantages. The research in \cite{9256688} developed an optimization technique for a micro aerial vehicle (MAV) follower for a UGV. They demonstrated the creation of augmented 2D maps using MAV-UGV coordination. Such a coordinated formation yielded detailed maps from top-view angles captured by the MAV and from side-view angles by the UGV.
Beneficial traits among heterogeneous agents often manifest in sensing, mobility, and communication roles. By combining agents with overlapping yet distinct capabilities, the swarm can achieve performance levels that are unattainable by any single agent type. This synergy is essential in missions that demand both global situational awareness and fine-grained local interaction.

\begin{figure*}
    \centering
    \includegraphics[width=0.8\linewidth]{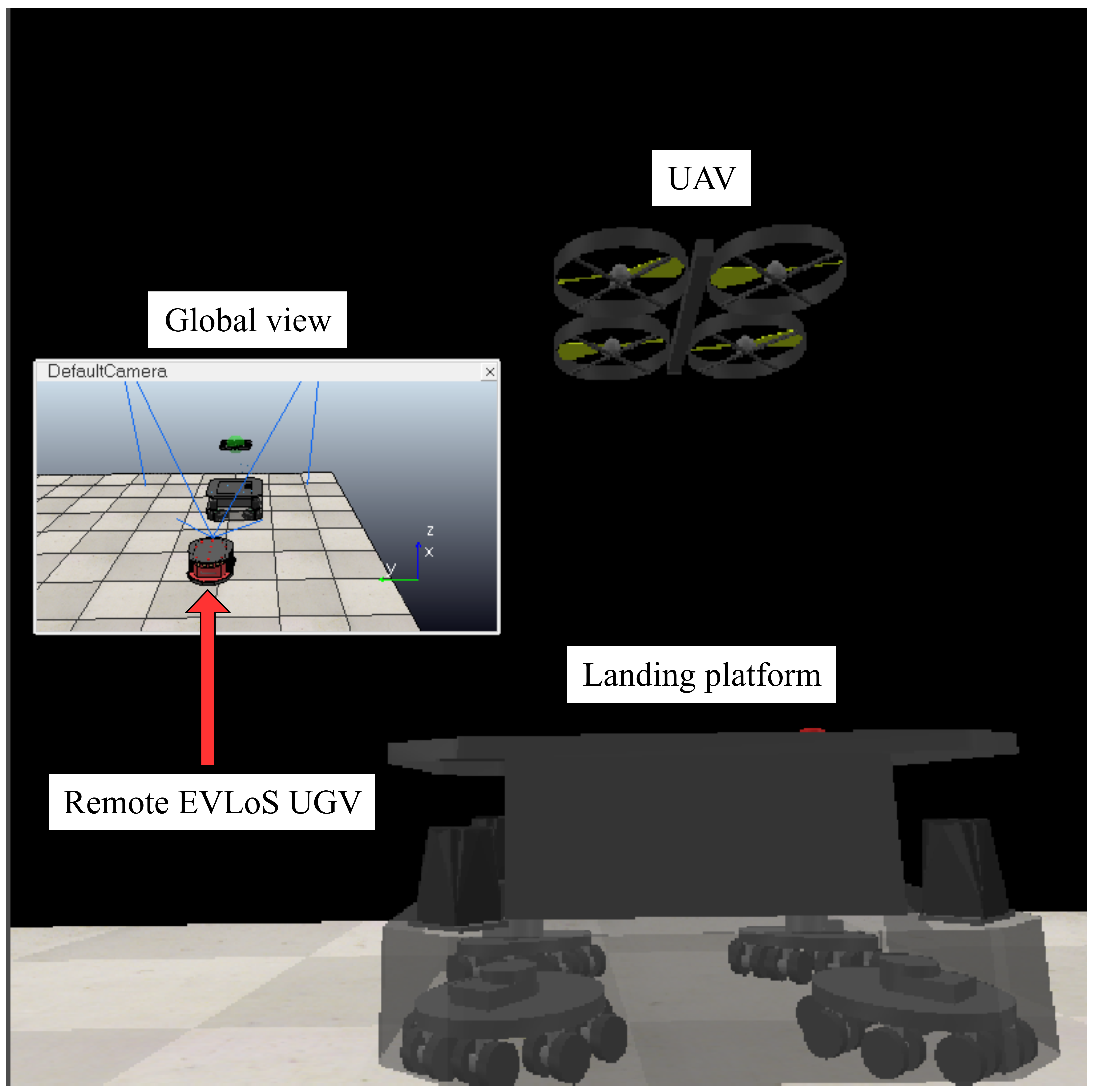}
    \caption{A heterogeneous robot combination working to establish EVLOS}
    \label{fig:evlos}
\end{figure*}

An alternative to BVLOS (Beyond Visual Line of Sight) operations is EVLOS (Extended Visual Line of Sight), in which an observer is closer to the aerial vehicle and directly communicates with the pilot, who is not near the vehicle. EVLOS has certain advantages over BVLOS, including being easier and faster to implement, requiring a simpler setup, providing a streamlined "stepping stone" to BVLOS, and allowing for wider coverage than VLOS (Visual Line of Sight). It also offers greater operational flexibility for tasks such as linear inspections. It's more agile and reduces logistics for large-area jobs, even with BVLOS's higher regulatory hurdles. However, conventional EVLOS techniques use a human observer. Advanced integration of multi-capability vehicles into a single swarm enables a remote-operated UGV to track the aerial vehicle and provide the pilot with a video feed of the mission. An additional landing platform UGV is present in the scene and is also tracked, as it provides charging and landing capacity for the UAV. This successful simulation, shown in Figure \ref{fig:evlos}, is a proof of concept for the potential to extend functionality through heterogeneity. Decentralized coordination mechanisms, enabled by graph neural networks ~\cite{goarin2024graph}, allow these complementary teams to switch roles without an active central authority. The mission carries on despite intermittent communication links or limited bandwidth. The article~\cite{https://doi.org/10.1155/2024/6643424} shows that adaptive replanning strategies are valuable for dynamic target-search missions. In such missions, faster-moving scouts locate targets and send relevant, relative information to a strike or inspection platform with a defined set of capabilities. This mechanism was found to reduce overall mission time in a simulated urban search campaign by 18-25\%.

A simple, low-level experiment described below shows the effects of adding an assistant UGV to a mission previously conducted by a single UAV on map coverage and knowledge map quality.

\begin{figure*}
    \centering
    \includegraphics[width=1\linewidth]{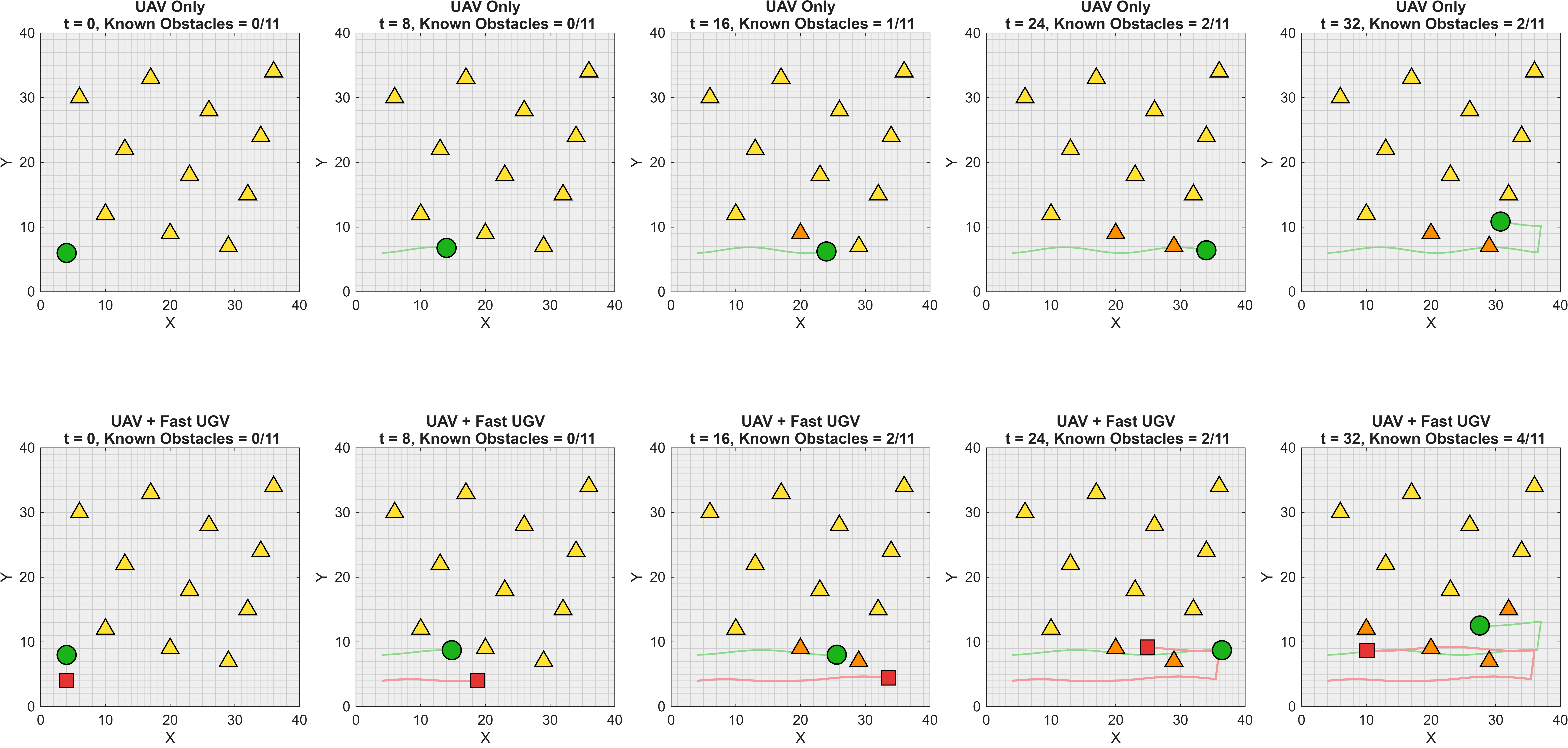}

    \caption{Static time-series comparison of obstacle mapping in a top-down environment. In the top row, a UAV-only system explores the map and gradually identifies obstacles. In the bottom row, a cooperative UAV+UGV team performs the same task, with the faster ground vehicle covering additional lower and middle regions of the map.}
    
    \label{fig:obstacle_mapping_subplots}
\end{figure*}

\begin{figure}
    \centering
    \includegraphics[width=1\linewidth]{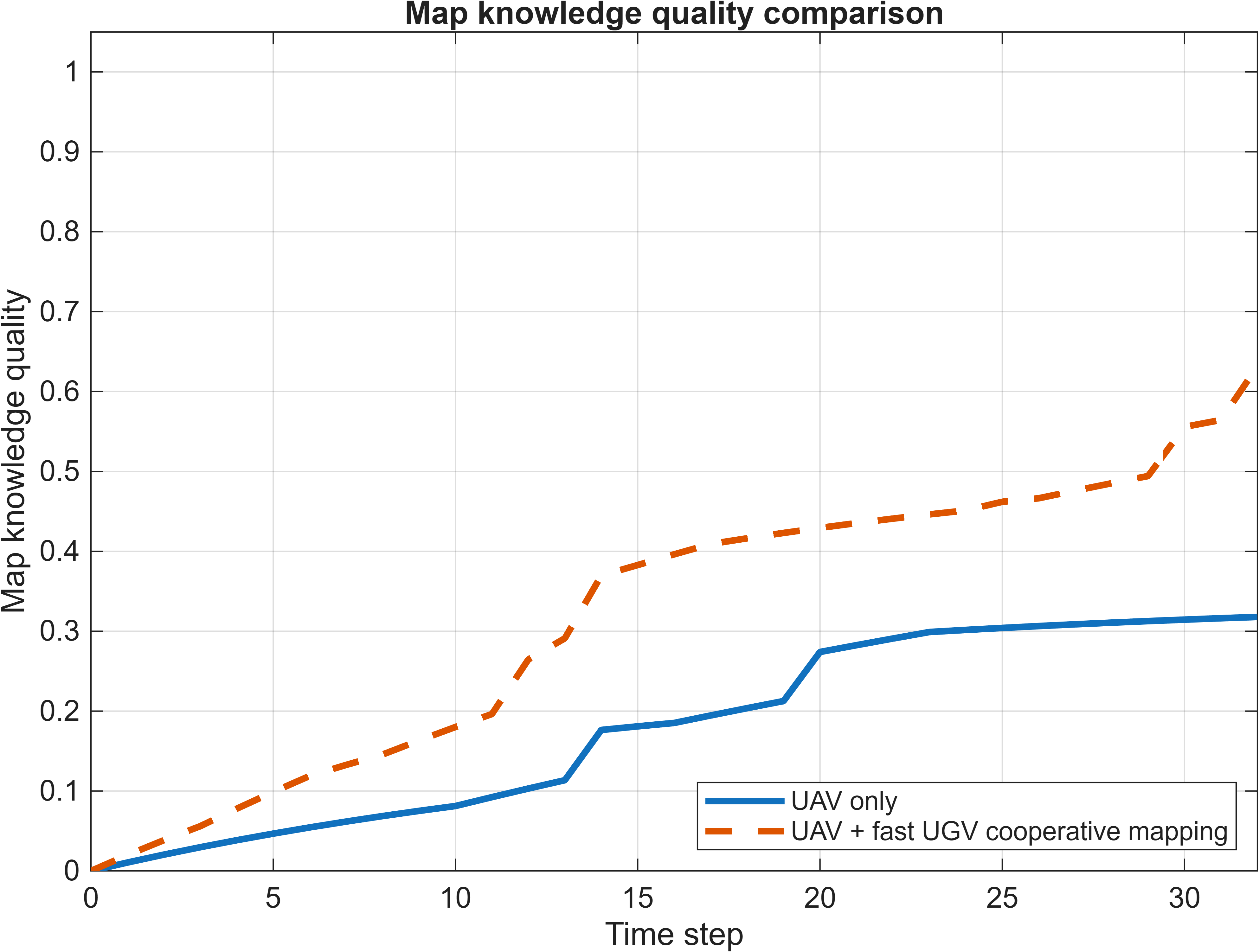}
    \caption{Map knowledge quality as a function of time for the UAV-only and cooperative UAV+UGV configurations. The UAV-only case exhibits slower growth and lower final map quality, whereas the cooperative case improves more rapidly due to complementary sensing and distributed exploration.}
    \label{fig:map_quality}
\end{figure}

\begin{equation}
\begin{split}
Q_j(t+1) = \min \Bigl(1,\; Q_j(t) &+ \alpha_u \,\mathbf{1}\!\left[\|\mathbf{o}_j-\mathbf{p}_u(t)\| \le r_u\right] \\
&+ \alpha_g \,\mathbf{1}\!\left[\|\mathbf{o}_j-\mathbf{p}_g(t)\| \le r_g\right]\Bigr),
\end{split}
\label{eq:obs_quality}
\end{equation}

\begin{equation}
\mathcal{M}(t)=\frac{1}{N_o}\sum_{j=1}^{N_o} Q_j(t),
\label{eq:map_quality}
\end{equation}

Let $Q_j(t)\in[0,1]$ denote the mapping confidence associated with obstacle $j$ at time step $t$. As shown in \eqref{eq:obs_quality}, this confidence increases whenever the obstacle lies within the sensing range of the UAV or the UGV. Here, $\mathbf{o}_j$ is the obstacle location, $\mathbf{p}_u(t)$ and $\mathbf{p}_g(t)$ are the UAV and UGV positions, and $r_u$ and $r_g$ denote their sensing radii. The parameters $\alpha_u$ and $\alpha_g$ represent the confidence increment produced by each sensing platform.

The global map knowledge quality is defined in \eqref{eq:map_quality} as the average obstacle knowledge over all $N_o$ obstacles. A higher value of $\mathcal{M}(t)$ indicates a more complete and reliable environmental representation. Because the cooperative UAV+UGV configuration observes obstacles from complementary viewpoints, it yields a faster increase in $\mathcal{M}(t)$ than the UAV-only configuration.

Figure~\ref{fig:obstacle_mapping_subplots} compares obstacle-mapping performance for a UAV-only system and a cooperative UAV+UGV system. The green circle denotes the UAV, the red square denotes the ground vehicle, yellow triangles denote unmapped obstacles, and orange triangles denote mapped obstacles. The cooperative configuration identifies obstacles more rapidly and yields stronger map knowledge growth. In the UAV-only case, obstacle discovery progresses gradually as the aerial platform alone sweeps the environment. In contrast, the cooperative case benefits from a faster-moving ground vehicle that covers additional lower and middle regions of the map in parallel with the UAV. As a result, obstacles are identified earlier and more consistently in the cooperative setting. This trend is also reflected in Figure~\ref{fig:map_quality}, where the cooperative configuration exhibits a steeper increase in map knowledge quality and reaches a higher final value than the UAV-only baseline.

The authors of \cite{10.3389/frobt.2021.616950} proposed a competitive coevolutionary genetic algorithm to create a multi-space heterogeneous swarm comprising land, water-surface, and ground vehicles to detect escaping targets. Their system achieved a higher detection rate than other algorithms, both in terms of the area covered and in its ability to detect early. They attribute this performance increase to the swarm agents' varying proximity radii and operating environments. They cite earlier literature \cite{10.1007/978-3-319-99582-3_22} in which UGVs, with high levels of interaction and autonomy, complement UAVs by providing communication support and the added advantage of an altitude perspective.

\subsubsection{Increase in Performance}
A hetero-UAV swarm system, introduced in \cite{6731679}, featured a lead agent for smaller teams of UAVs to enter hostile territory and search for a target. Their super-agent-led swarm exhibited improved target-search performance in hostile terrain. A similar marsupial cooperative team, comprising a UAV and a UWSV, was used in a study collaboration with the Portuguese Navy by \cite{7761074} to search for shipwrecked sailors. The authors stated that using such teams improved their ability to patrol the sea, search for shipwreck survivors, and provide basic life support until emergency services arrived. The surface vehicle served as the UAV's takeoff and landing platform, and the water-surface vehicle could extend the UAV's search range. The increase in heterogeneous swarm performance is not limited to SAR (Search and Rescue).
Application-specific improvements are also observed.
Figure \ref{fig:FAA-hetero} shows a heterogeneous vehicle swarm working together to examine the external structure of an aircraft for defects. The UAV covers the upper fuselage, whereas the ground robot covers the underside in sections where the UAV might have difficulty navigating or lose signal coverage.

The effects of adding unique agents assigned special functions to a previously homogeneous swarm can be observed in the experiment below. 

\begin{equation}
\mathcal{V}_c(t)=\left\{ i \in \mathcal{V} \; \middle| \; \exists \text{ a communication path from } i \text{ to } g \right\},
\label{eq:connected_set}
\end{equation}

\begin{equation}
\rho(t)=\max_{i \in \mathcal{V}_c(t)} \left\| \mathbf{p}_i(t)-\mathbf{p}_g \right\|,
\label{eq:mission_reach}
\end{equation}

\begin{equation}
(i,j)\in\mathcal{E}(t) \iff \|\mathbf{p}_i(t)-\mathbf{p}_j(t)\| \le R_{ij},
\label{eq:comm_edge}
\end{equation}

Let $\mathcal{V}$ denote the set of UAVs and let $g$ denote the ground control node. At each time step, the communication graph is formed by the set of links that satisfy the communication-range constraint in \eqref{eq:comm_edge}. A UAV is considered operationally connected if there exists a multi-hop communication path from that UAV to ground control. The set of such UAVs is given by $\mathcal{V}_c(t)$ in \eqref{eq:connected_set}.

The communication-supported mission reach is defined in \eqref{eq:mission_reach} as the maximum distance between ground control and any UAV that remains connected to the network. In the direct-communication case, $\rho(t)$ is limited by the ground-control communication radius. In the heterogeneous case, relay UAVs create additional aerial links, allowing the swarm to extend farther while maintaining end-to-end connectivity.

\begin{figure*}
    \centering
    \includegraphics[width=0.95\linewidth]{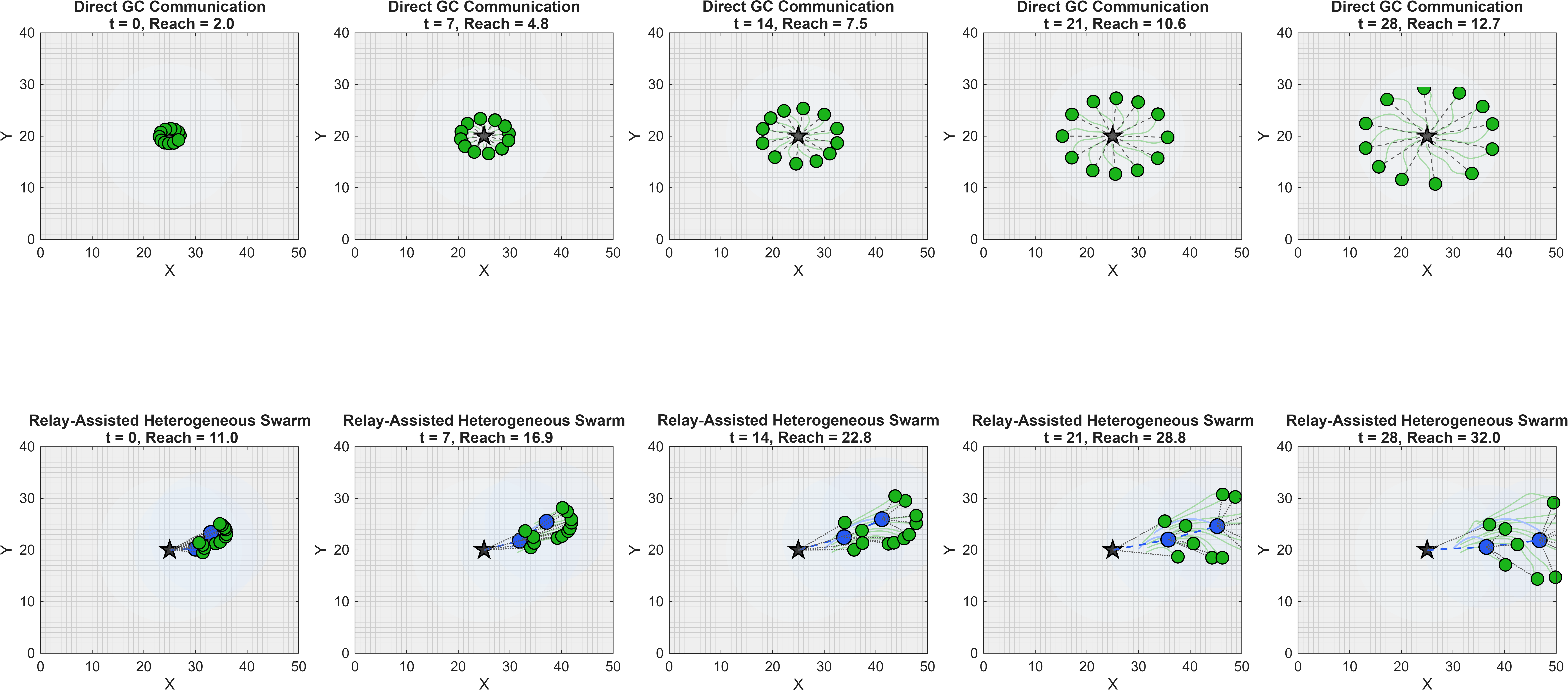}
    \caption{Static time-series comparison of communication-constrained swarm deployment. In the top row, all UAVs communicate directly with ground control, which limits how far the swarm can move from the base station. In the bottom row, two relay UAVs (blue) support multi-hop communication between the swarm and ground control, enabling the remaining UAVs (green) to extend farther into the mission space while preserving connectivity.}
    \label{fig:relay_comm_subplots}
\end{figure*}

\begin{figure*}[t!]
    \centering
    \includegraphics[width=0.95\linewidth]{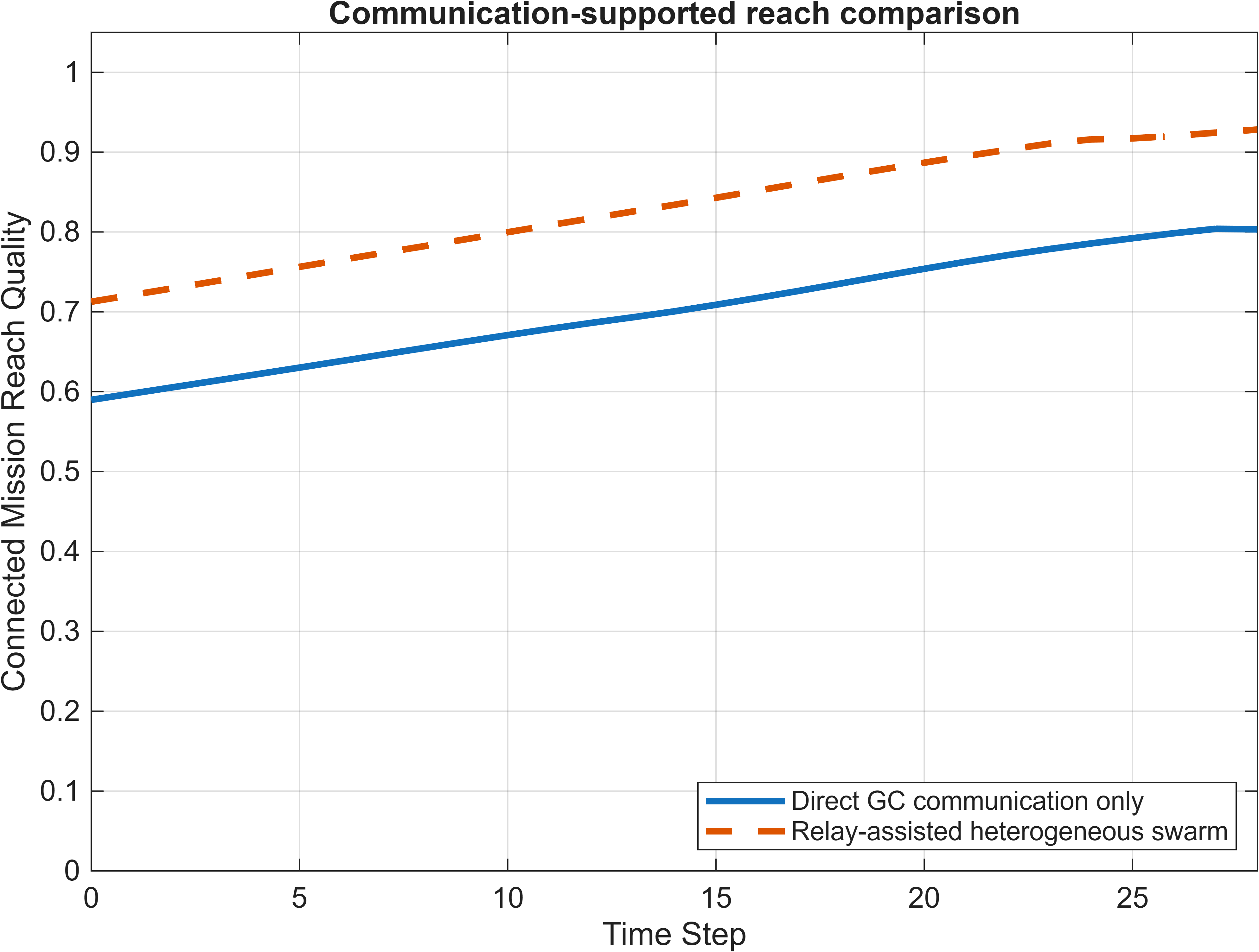}
    \caption{Connected mission reach quality over time for the direct-communication swarm and the relay-assisted heterogeneous swarm. The relay-assisted configuration achieves greater communication-supported reach because the relay UAVs extend the effective connectivity region beyond the direct ground-control range.}
    \label{fig:relay_comm_quality}
\end{figure*}

Figure~\ref{fig:relay_comm_subplots} illustrates the effect of relay-assisted communication on swarm deployment range. In the direct-communication case, all UAVs must remain within the ground control's communication footprint, which constrains the swarm to operate near the base station. In contrast, the heterogeneous configuration introduces two relay UAVs that maintain communication links with each other and with ground control, thereby extending the effective network boundary. As a result, the remaining UAVs can move farther into the mission space while remaining connected. This trend is further confirmed in Figure~\ref{fig:relay_comm_quality}, where the relay-assisted swarm achieves a higher connected mission reach quality over time.

\begin{figure*}[t!]
    \centering
    \includegraphics[width=0.91\linewidth]{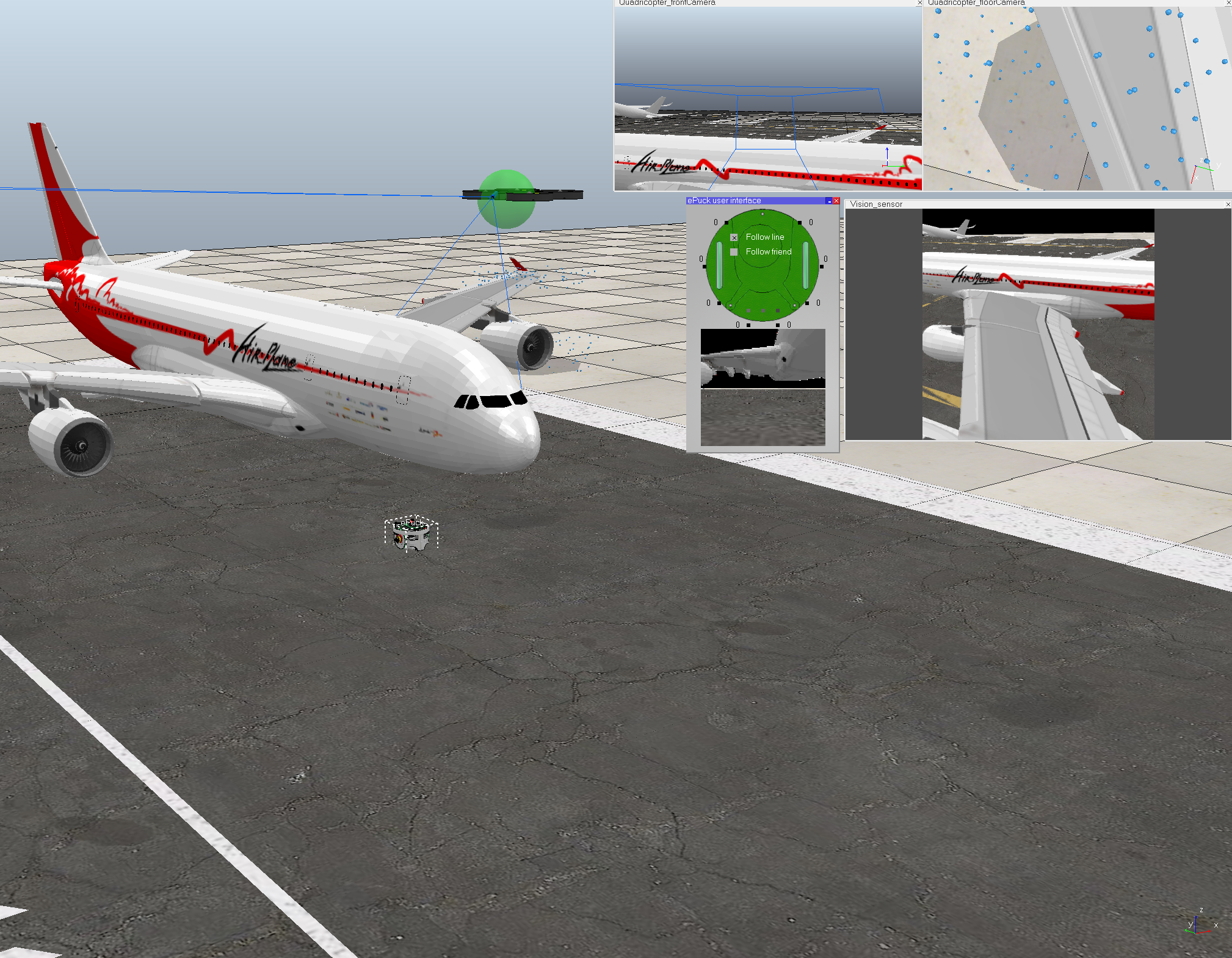}
    \caption{A UGV and UAV working together to perform a structural integrity check on an aircraft}
    \label{fig:FAA-hetero}
\end{figure*}
Agents in \cite{1501630} had different tolerances of collision and avoidance of other swarm members. They controlled a parameter, ``alpha,'' and found that modifying its value for agents altered their swarming nature and flight speed. Teams of heterogeneous agents with different alpha values achieved better overall performance in chemical cloud detection than homogeneous teams of agents with the same alpha value.
Performance gains in heterogeneous swarms are frequently nonlinear. The addition of specialized agents can yield disproportionate improvements in mission outcomes. This effect underscores the importance of strategic heterogeneity in design, rather than simply increasing swarm size or agent redundancy. Validation based on competitions provides objective performance comparisons. The CERBERUS heterogeneous aerial system~\cite{tranzatto2022cerberusautonomousleggedaerial} won the DARPA Subterranean challenge by using its unique mobility and sensing abilities to complete artifact detection tasks in situations where teams with only aerial or ground-based sensors failed because of terrain obstacles, communication blackouts, or sensor degradation in environments with dust and sparse light.

%%%%%%%%%%%%%%%%%%%%%

\begin{table*}[h]
\centering
\caption{Mission Characteristics to Heterogeneity Strategy Mapping}
\label{tab:design_guidelines}
\renewcommand{\arraystretch}{1.3}
\begin{tabular}{|p{0.90\columnwidth}|p{0.90\columnwidth}|}
\hline
\textbf{Mission Characteristic} & \textbf{Recommended Heterogeneity Type} \\
\hline
Uniform environment, dynamic roles & Nature-based \\
\hline
Hardware replacement speed critical & Nature-based \\
\hline
Multi-environment coverage required & Operational-space + Hardware \\
\hline
Sensing diversity needed & Hardware-based \\
\hline
Strict energy/logistics constraints & Nature-based or single-space hardware \\
\hline
Large geographic area, communication relay needed & Hardware-based (fixed-wing + rotorcraft) \\
\hline
Adversarial environment, jamming risk & Nature-based (learned, decentralized) \\
\hline
\end{tabular}
\end{table*}

%%%%%%%%%%%%%%%%%%%%%%%

The operational factors and resilience benefits examined in this section indicate that heterogeneity makes implementation more challenging but also improves performance. Researchers continue to address issues in communication, energy management, and control coordination. Still, documented deployments, ranging from lab experiments to competition settings, demonstrate that heterogeneous swarm systems can operate in real-world settings. The next section draws on these results to outline concluding statements and future areas of research that could help make heterogeneous swarm technology ready for real-world use.

\clearpage          
\onecolumn 
\section{Overview of Current Work in Heterogeneous Robotic Teams}
\label{overview}
% Required in preamble: \usepackage{longtable, array}
% Place this block directly in your .tex file where you want the table to appear.
% The \onecolumn ... \twocolumn switching handles two-column document layouts.

\small
\renewcommand{\arraystretch}{1.3}
\begin{longtable}{|p{2.2cm}|p{14.5cm}|}
\caption{Summary of Heterogeneous Multi-Robot Swarm Studies}
\label{tab:swarm_studies} \\

\hline
\textbf{Reference} & \textbf{Study Contribution} \\
\hline
\endfirsthead

\multicolumn{2}{c}{\tablename\ \thetable{} -- \textit{Continued from previous page}} \\
\hline
\textbf{Reference} & \textbf{Study Contribution} \\
\hline
\endhead

\hline
\multicolumn{2}{r}{\textit{Continued on next page}} \\
\endfoot

\hline
\endlastfoot

\cite{6731679} & Proposes a decentralized heterogeneous UAV swarm using APF and ACO with a Super-Agent class for target recognition in hostile airspace. Leader-led heterogeneous swarms achieve higher search rates and fewer unconfirmed targets than homogeneous swarms. \\
\hline
\cite{9389145} & Investigates heterogeneous maritime swarms of high- and low-speed vehicles using a PSO-based adaptive neighbor topology for moving-target tracking. High-speed agents improve reacquisition speed but reduce optimal swarm connectivity. \\
\hline
\cite{1501630} & Demonstrates that heterogeneous UAV teams outperform homogeneous swarms in chemical plume detection, with fast UAVs forming an outer layer and slow UAVs penetrating inward for better coverage and tracking. \\
\hline
\cite{Makkapati2020ApolloniusAA} & Introduces the Apollonius Allocation (A2) algorithm for coordinating heterogeneous UAV pursuers to capture multiple evaders using dynamic divide-and-conquer task allocation. Guarantees finite-time capture and real-time agent reassignment under heterogeneous speed conditions. \\
\hline
\cite{deka2020naturalemergenceheterogeneousstrategies} & Presents FortAttack, a mixed cooperative-competitive environment where GNN agents trained via PPO develop emergent heterogeneous behaviors such as deception, coordination, and role specialization. Demonstrates that heterogeneous strategies can arise naturally among architecturally homogeneous agents. \\
\hline
\cite{8833293} & Proposes an Improved Multi-objective Quantum-behaved PSO (MOQPSO) with GA-inspired operators for mission planning in heterogeneous anti-radar UAV formations. Yields more stable and diverse Pareto-optimal trade-offs than standard MOPSO across time, cost, and reward objectives. \\
\hline
\cite{deng2013cooperative} & Formulates heterogeneous UAV task assignment as a constrained combinatorial problem and proposes a Modified Genetic Algorithm (MGA) with multi-type genes. Demonstrates faster convergence and more efficient workload distribution than standard GA and PSO. \\
\hline
\cite{9215994} & Evaluates three UAV behavioral types (Social Searcher, Antisocial Searcher, Relay) across 1,000 simulated post-disaster search-and-rescue experiments. Finds that behavioral diversity enables detection of over 90\% of survivors within 40 minutes, surpassing homogeneous designs. \\
\hline
\cite{7271492} & Develops a multi-domain AUV-USV-UAV system for real-time oil spill monitoring by fusing acoustic and visual data across platforms. Field trials in the Adriatic Sea confirm the effectiveness of cross-domain coordination for marine environmental surveillance. \\
\hline
\cite{JU2022107336} & Reviews heterogeneous agricultural multi-robot systems integrating UAVs, ground robots, and manipulators for monitoring, spraying, and harvesting. Reports 20--40\% labor cost reductions and 30--50\% pesticide savings, while identifying open challenges in GPS-denied navigation and farmer adoption. \\
\hline
\cite{app10051583} & Develops an autonomous UAV-USV marsupial system with arresting cable and wave compensation for fixed-wing UAV recovery at sea. Achieves over 80\% recovery success within 2 minutes, enabling continuous maritime monitoring with minimal personnel. \\
\hline
\cite{7761074} & Proposes a collaborative UAV-USV team for maritime search and rescue, combining UAV visual-saliency aerial surveillance with USV thermal scanning and kit delivery. Navy trials confirm reduced response times and lower operational costs through heterogeneous robotic coordination. \\
\hline
\cite{6107312} & Introduces CARACaS, a flexible autonomy architecture enabling heterogeneous ASV/AUV teams with behavior-based control, real-time hazard avoidance, and adaptive mission planning. Demonstrated in coastal and riverine environments, supporting autonomous collaboration and resilience under dynamic conditions. \\
\hline
\cite{10.1007/s10846-009-9334-x} & Presents HUAS, integrating miniature and small UAVs over a multi-tier wireless network with dynamic publish-subscribe data exchange and mothership coordination. Hardware-in-the-loop experiments validate scalable net-centric C3 for autonomous convoy tracking and collaborative target assignment. \\
\hline
\cite{7849525} & Describes ASIMUT, a multi-level heterogeneous UAV swarm architecture for distributed surveillance combining chaos-based mobility, high-level data fusion, and an Object-Oriented World Model. Demonstrates real-time target tracking, autonomous re-tasking, and resilient multi-tier communication for improved operator situational awareness. \\
\hline
\cite{tranzatto2022cerberusautonomousleggedaerial} & Presents CERBERUS, a system-of-systems combining quadrupeds, quadrotors, and rovers for autonomous underground exploration and artifact detection. Achieves heterogeneous coordination with role-switching and resilient multi-modal SLAM under adverse conditions in the DARPA Subterranean Challenge. \\
\hline
\cite{s20185026} & Proposes a multi-swarm fruit fly optimization algorithm (MFOA) for cooperative mission planning of heterogeneous UAVs incorporating 3D path planning and realistic kinematic constraints. Outperforms standard FOA in convergence stability and scalability across small-, medium-, and large-scale simulations. \\
\hline
\cite{8995977} & Extends the Consensus-Based Bundle Algorithm (CBBA) to heterogeneous UAV teams with capacity, load, and endurance constraints via a resource-demand matrix and time-discounted reward model. Achieves faster convergence and higher task-completion rates than standard CBBA while scaling efficiently. \\
\hline
\cite{8491531} & Develops a fuzzy sliding-mode consensus formation controller for heterogeneous Mecanum-wheeled robots under dynamic uncertainties and external perturbations. Achieves finite-time convergence and robust trajectory tracking without requiring accurate model knowledge. \\
\hline
\cite{10.3389/frobt.2021.616950} & Presents CROMM-MS, a chaotic R\"{o}ssler mobility model for heterogeneous UAV-UGV-UMV perimeter surveillance co-optimized with intruder strategies via a Competitive Coevolutionary GA. Achieves approximately 40\% higher detection accuracy than the baseline across four multi-intruder case studies. \\
\hline
\cite{app11199145} & Introduces an L1 adaptive control framework for heterogeneous UAV-AUV formation and containment with actuator delay constraints, using DDS middleware and potential-field obstacle avoidance. Demonstrates formation integrity and cross-domain interoperability in 2D and 3D simulations. \\
\hline
\cite{9237985} & Develops a PSO-based multi-objective mission planning framework for heterogeneous UAV formations in mountainous terrain, addressing threat avoidance, terrain masking, and sensor coverage. Simulations show enhanced flight safety and mission efficiency over baseline planners in complex 3D environments. \\
\hline
\cite{9264359} & Proposes a distributed formation-tracking controller for heterogeneous UAV-UGV swarms under switching directed communication topologies, using a distributed observer and algebraic graph theory. Piecewise Lyapunov analysis confirms stability and resilience to convergence under communication disruptions. \\
\hline
\cite{10301527} & Presents a MacDec-POMDP framework with macro actions for asynchronous heterogeneous multi-robot coordination, combined with a Qrainbow value-mixing network for decentralized execution. Achieves 8--16\% time savings over baselines on cooperative foraging and search-and-rescue tasks while reducing inter-robot collisions. \\
\hline
\cite{rs12101608} & Proposes an integrated 3D registration and segmentation pipeline for heterogeneous aerial and ground robots using feature-based point cloud registration and probabilistic data fusion. Experimental results demonstrate robust scene understanding enabling coordinated perception across heterogeneous platforms. \\
\hline
\cite{chang2022lamp} & Presents LAMP 2.0, a centralized multi-robot LiDAR SLAM system for large-scale underground search and rescue with adaptive loop closure and outlier-resistant pose graph optimization. Achieves low trajectory drift over multi-kilometer traversals across mines, caves, and industrial environments. \\
\hline
\cite{10.5555/2484920.2485103} & Introduces an integer programming model for coordinated UGV-MAV indoor exploration that jointly optimizes frontier allocation with visibility and inter-agent coordination constraints. Outperforms incremental frontier methods in travel distance and mapping quality on real Pioneer P3DX and Parrot AR.Drone platforms. \\
\hline
\cite{goarin2024graph} & Proposes DGNN-GA, a heterogeneous graph neural network for decentralized goal assignment that mimics the centralized Hungarian algorithm under limited communication. Outperforms auction-based baselines with limited communication rounds and generalizes to larger teams than seen during training. \\
\hline
\cite{9256688} & Presents HD-RHC, a receding horizon control architecture for heterogeneous UAV swarms balancing search coverage, mission point completion, and energy efficiency via Monte Carlo and simulated annealing cost weighting. AirSim simulations confirm scalability across large fleets with varying payload sizes. \\
\hline
\cite{jmse9111314} & Implements a leader-follower consensus scheme for heterogeneous UAV-USV systems using APF and fuzzy sliding mode control, with the UAV providing vision-based reference for USV formation. Demonstrates finite-time convergence, collision avoidance, and field-of-view maintenance under dynamic sea states. \\
\hline
\cite{horyna2023decentralized} & Demonstrates communication-free UAV swarm search and rescue using bio-inspired flocking and L\'{e}vy flights, achieving 77\% victim detection in simulated disaster zones. Physical quadrotor field experiments confirm victim localization without reliance on inter-agent communication networks. \\
\hline
\cite{deng2023distributed} & Proposes a two-level distributed scheduling framework for heterogeneous UAV surveillance and attack missions using a modified contract net protocol and predictive management. Teams of 6--15 UAVs complete operations 15\% faster than solo approaches, with scalability confirmed across team sizes. \\
\hline
\cite{9508420} & Introduces Neural-Swarm2, a learning-based framework that models aerodynamic downwash and ground effects among heterogeneous multirotor configurations using neural networks. Hardware tests with up to 10 vehicles achieve path-tracking errors below 0.15~m and outperform model-based approaches by 40\%. \\
\hline
\cite{WANG2024315} & Proposes a hierarchical air-ground coordinated swarm framework where aerial vehicles provide situational awareness and ground vehicles perform manipulation tasks. Warehouse simulations show a 25\% reduction in task completion time compared to ground-only logistics systems. \\
\hline
\cite{ai5020029} & Develops a transformer-based attention mechanism for adaptive heterogeneous robot teaming that dynamically weights team members by capability and task requirements. Mixed manipulator-mobile-aerial teams complete tasks 30\% faster than role-assigned counterparts across varied mission types. \\
\hline
\cite{BARTH2024689} & Investigates deep reinforcement learning for heterogeneous rover-lander planetary exploration, training in simulated Martian environments with dust storms and communication blackouts. The learned policy discovers novel coordination strategies that use landers as navigation beacons for rovers under adverse conditions. \\
\hline

\end{longtable}
\normalsize
\clearpage
\twocolumn

\section{Mapping heterogeneity to resilience}
\label{map}
This study presents evidence that heterogeneous swarms exhibit greater resilience than homogeneous swarms. This section maps specific, commonly encountered failure modes to the heterogeneity types that most effectively mitigate them. This analysis draws on deployments reviewed in section \ref{overview} and provides a structured basis for resilience-oriented design. Table \ref{tab:design_guidelines} provides an overview of various mission characteristics and the recommended heterogeneity needed in the swarm for optimal performance.
\subsection{Agent loss}
The loss of individual agents due to hardware failure, hostile action, or environmental damage is the most commonly studied failure mode in the swarm literature. Homogeneous swarms respond to agent loss along narrow capability reductions. Fewer agents mean reduced coverage or slower task completion, but the swarm's functional profile remains largely unchanged. Heterogeneous swarms face a more complex failure dynamic, since losing a specialized agent can eliminate an entire capability class. However, the reviewed literature suggests two mechanisms for mitigation. First, nature-based heterogeneity enables dynamic role reassignment: if a designated relay agent fails, another agent can assume the relay role without hardware changes, as demonstrated in \cite{1501630} and \cite{deka2020naturalemergenceheterogeneousstrategies}. Second, hardware redundancy across platforms means that alternative agent types can partially replicate some capabilities. In \cite{tranzatto2022cerberusautonomousleggedaerial}, the loss of aerial scouts during the DARPA Subterranean Challenge was partially mitigated by quadruped robots equipped with alternative sensing modalities, demonstrating that cross-platform redundancy can sustain mission progress under agent-loss conditions.
\subsection{Communication failure and jamming}
Communication degradation is particularly consequential in heterogeneous swarms because agents may use different protocols, and a relay node failure can partition the swarm. The reviewed literature identifies several heterogeneity-enabled mitigations. Architecturally, assigning dedicated relay roles to high-endurance, high-altitude platforms as in \cite{7849525} and \cite{10.1007/s10846-009-9334-x} creates a communication hierarchy that is more robust than flat peer-to-peer topologies. Behaviorally, learned coordination policies trained under simulated jamming conditions have demonstrated resilience to communication attacks, with one study reporting a 40\% higher mission success rate under jamming conditions than under traditional protocols \cite{10107729}. Decentralized approaches, such as the communication-free swarm coordination demonstrated in \cite{horyna2023decentralized}, represent an extreme yet effective mitigation strategy. By eliminating dependence on inter-agent communication, these systems are inherently immune to jamming at the cost of reduced coordination efficiency.
\subsection{GPS denial}
Operations in GPS-denied environments, such as underground, indoor, or contested airspace, are increasingly required. Heterogeneous swarms address this challenge more effectively than homogeneous systems, as diverse sensor suites can serve as substitutes for GPS-based localization. The LAMP 2.0 system \cite{chang2022lamp} demonstrated that multi-robot LiDAR-based SLAM can maintain localization accuracy across kilometer-scale underground traversals. The CERBERUS system \cite{tranzatto2022cerberusautonomousleggedaerial} combined aerial and ground platforms with cameras, LiDAR, and inertial sensors, achieving robust localization that would have failed with any single sensor. The key design principle is sensor complementarity: hardware heterogeneity that provides overlapping yet distinct sensing capabilities yields a localization system more robust than the sum of its parts.
\subsection{Sensor degradation}
Environmental conditions such as dust, smoke, low light, and water selectively degrade specific sensor types. Heterogeneous swarms incorporating multiple sensors are inherently more robust to selective sensor degradation than homogeneous systems relying on a single sensing approach. In \cite{tranzatto2022cerberusautonomousleggedaerial}, dust accumulation in underground tunnels degraded visual sensors but left LiDAR performance largely intact; teams relying solely on visual sensing failed tasks that multi-modal teams completed. This pattern generalizes: thermal imaging supplements visual sensors in low-light conditions, acoustic sensing substitutes for visual sensors underwater, and LiDAR maintains performance in dusty or smoky environments where cameras fail.
\subsection{Energy depletion}
Energy constraints affect aerial platforms much more as they consume power at higher rates than ground vehicles. Heterogeneous swarms that incorporate ground or surface vehicles as mobile charging platforms, as demonstrated in \cite{app10051583} and \cite{JU2022107336}, extend mission endurance beyond that of aerial-only swarms. This represents a structural resilience benefit: the energy failure mode that would terminate a homogeneous aerial swarm becomes a manageable logistics problem in a multi-domain system with surface-based support assets.

\begin{table*}[h]
\centering
\caption{Failure Mode to Heterogeneity Mitigation Mapping}
\label{tab:failure_modes}
\renewcommand{\arraystretch}{1.3}
\begin{tabular}{|p{0.15\textwidth}|p{0.2\textwidth}|p{0.3\textwidth}|p{0.15\textwidth}|}
\hline
\textbf{Failure Mode} & \textbf{Most Vulnerable Swarm Type} & \textbf{Heterogeneity Mitigation} & \textbf{Key References} \\
\hline
Agent loss & Hardware-specialized swarms & Dynamic role reassignment (nature-based) & \cite{1501630}, \cite{tranzatto2022cerberusautonomousleggedaerial}, \cite{8995977} \\
\hline
Communication jamming & Centralized, flat topologies & Relay hierarchy; learned jamming-resistant policies & \cite{7849525}, \cite{horyna2023decentralized}, \cite{10107729} \\
\hline
GPS denial & Vision/GPS-dependent systems & Multi-modal SLAM; sensor-diverse hardware & \cite{tranzatto2022cerberusautonomousleggedaerial}, \cite{chang2022lamp}, \cite{8598942} \\
\hline
Sensor degradation & Single-modality systems & Cross-modal sensing redundancy & \cite{tranzatto2022cerberusautonomousleggedaerial}, \cite{chang2022lamp}, \cite{drones6040094}
\\
\hline
Energy depletion & Aerial-only swarms & Surface/ground charging platforms & \cite{JU2022107336}, \cite{app10051583}, \cite{10510424}, \cite{https://doi.org/10.1002/rob.21856}\\
\hline
\end{tabular}
\end{table*}

\section{Conclusion}
\label{conclude}

The consideration of heterogeneous agents in swarms is gaining momentum due to the need for enhanced swarm capabilities across diverse applications. The inclusion of agents with varying specialties and operational domains has been found to overcome mission output barriers arising from homogeneous-agent limitations. Recent developments from 2022 to 2024 have shown significant progress in practical applications, such as validated systems for underground exploration~\cite{chang2022lamp,tranzatto2022cerberusautonomousleggedaerial}, agricultural automation~\cite{JU2022107336}, learning-based decentralized coordination~\cite{goarin2024graph}, and asynchronous multi-agent control~\cite{10301527}. This shows that heterogeneous swarm systems are moving from the lab to the field. Table \ref{tab:failure_modes} presents key failures observed in swarms in environments, the swarm types most vulnerable to these failures, and suggested mitigation strategies.

The contributions of this study are threefold. 
\begin{itemize}
    \item This study specifies three classification categories for heterogeneous swarms based on nature, structure, and operational space. 
    \item It examines the operational complexities to be considered when implementing a heterogeneous swarm. While it is entirely achievable for homogeneous agents to have different behaviors, they are, in turn, limited by their hardware. For example, a fixed-wing aircraft that requires more space for takeoff and landing cannot perform vertical takeoff and landing (VTOL) operations. Similarly, bulky aircraft carrying heavier sensor arrays or payloads cannot effectively function as ``fast'' agents in a swarm. Such functions require lighter, more agile agents that lack features such as greater onboard computational power or personal defense systems, but can conduct quick reconnaissance sorties to provide environmental data to the swarm. Therefore, it is highly likely that heterogeneous swarms will include both behavioral differences and structured agents, as highlighted in Table \ref{tab:classification}. Creating labels for them is deemed necessary and was under-discussed in previous research.
    \item The potential benefits of increased operational performance and resilience through the inclusion of heterogeneous agents in swarms are discussed. While the inclusion of heterogeneous agents is inherently more complex to operate effectively, the benefits are significant. Recent deployments have shown that asynchronous coordination can significantly reduce mission completion time. Similarly, learning-based attention mechanisms can reduce collision rates, and agricultural applications can save on labor costs. Additionally, artifact detection has been successful in competition environments where homogeneous teams failed. This study provides a brief examination of the benefits of including heterogeneous agents in UAV swarms, demonstrating performance that exceeds that of simple-agent UAV swarms.
\end{itemize}

As uncrewed vehicle swarms transition from controlled experimental settings to real-world deployments, heterogeneity will likely become a defining characteristic rather than an optional enhancement. Future swarm systems are expected to incorporate adaptive role assignment, cross-domain coordination, and intelligent resource management as standard features. Further research is needed to establish standardized evaluation metrics, develop interoperable control architectures, and address emerging challenges, including cybersecurity and ethical deployment. The DARPA Subterranean Challenge and other competition-based benchmarks have demonstrated the utility of standardized evaluation environments for comparing different architectures. Learning-based methods also show promise for adaptive coordination that operates across diverse team compositions and mission contexts. Some important open problems include transferring learned policies from simulation to real-world settings, planning for long missions while accounting for energy use, designing interfaces for human-swarm interaction across teams with different skills, and applying formal verification methods to safety-critical deployments in contested environments. By providing a structured classification and synthesizing existing research, this study contributes to the foundational understanding required to design resilient, scalable, and mission-ready heterogeneous swarms. The classification framework, along with documented evidence of performance gains and operational deployments, provides researchers and practitioners with useful guidance for designing heterogeneous systems that balance complexity and capability improvements.

Recent research trends point to several promising paths for developing heterogeneous swarms. Methods are moving from simulation testing to real hardware testing. These methods can handle asynchronous execution~\cite{10301527}, communication constraints~\cite{goarin2024graph}, and behavioral specialization without explicitly assigning roles. Second, multi-modal sensor fusion is improving the perception and localization abilities of multi-robot operations that do not use GPS~\cite{chang2022lamp,tranzatto2022cerberusautonomousleggedaerial}. This enables robots to navigate underground, indoor, and cluttered environments. Third, "domain-specific deployments" in agriculture~\cite{JU2022107336} and other uses are proving profitable, with reports of lower costs and higher efficiency, making them more appealing to a wider audience. Fourth, standardized benchmarks and competitions help researchers compare their performance fairly and accelerate progress by involving the community in judging. These trends indicate that heterogeneous swarm technology is approaching the maturity level required for use in high-value applications.

\section*{Acknowledgment}

This paper is an extended version of the original conference paper accepted and presented to the 20th International Conference on Embedded Systems, Cyber-Physical Systems, \& Applications, held in \textbf{May 2022}. As of \textbf{March 28, 2026}, the conference proceedings have not been published. The original authors of the conference paper [AP, FAM, and TC] would like to thank Abhishek Joshi for updating, improving, and extending the original unpublished conference paper. This extended version has 60\% additional content added to the conference version overall, with multiple sections, such as sections II, V, and VI, being added in this version, along with the equations, experiments, and figures in  III and IV.

\section{AI USE DECLARATION}

The authors declare the use of the Grammarly writing assistant tool to improve the language and grammar of this manuscript. \textbf{No references were generated using AI}.

\bibliographystyle{IEEEtran} 
\bibliography{ref2}  

\end{document}